\documentclass{bmvc2k}

\usepackage{times}
\usepackage{epsfig}
\usepackage{graphicx}
\usepackage{amsmath}
\usepackage{amssymb}
\usepackage{pifont}
\usepackage{array}
\usepackage{mathtools}
\usepackage{multirow}
\usepackage{wrapfig}

\newcommand{\comment}[1]{}

\definecolor{rev_color}{rgb}{0.05, 0.58, 0.25}
\newcommand{\rev}[1]{#1}

\newcommand{\real}{{\rm I\!R}}
\newcommand{\myparagraph}[1]{\vspace{2pt}\noindent\textbf{#1}}
\newcommand{\set}{\mathcal}

\newcommand{\cmark}{\ding{51}}

\newcommand{\ours}{PIFS}
\newcommand{\oursFull}{Prototype-based Incremental Few-Shot Segmentation}
\newcommand{\oursFullAcro}{\textbf{P}rototype-based \textbf{I}ncremental \textbf{F}ew-Shot \textbf{S}egmentation}
\newcommand{\setting}{Incremental Few-Shot Segmentation}
\newcommand{\SET}{iFSS}
\newcolumntype{P}[1]{>{\centering\arraybackslash}p{#1}}

\newcommand{\figref}[1]{Fig.~\ref{#1}}
\newcommand{\tabref}[1]{Tab.~\ref{#1}}

\title{Prototype-based Incremental Few-Shot Semantic Segmentation}

\addauthor{Fabio Cermelli}{fabio.cermelli@polito.it}{1,2}
\addauthor{\vspace{-15pt}Massimiliano Mancini}{}{3}
\addauthor{\vspace{-15pt}Yongqin Xian}{}{4}
\addauthor{\vspace{-15pt}Zeynep Akata}{}{3}
\addauthor{\vspace{-15pt}Barbara Caputo}{}{1}

\addinstitution{
Politecnico di Torino, Italy
}
\addinstitution{
Italian Institute of Technology, Italy
}
\addinstitution{
University of Tübingen, Germany
}
\addinstitution{
ETH Zurich, Switzerland
}

\runninghead{Cermelli et al.}{Prototype-based Incremental Few-Shot S. Segmentation}

\def\eg{\emph{e.g}\bmvaOneDot} 
\def\ie{\emph{i.e}\bmvaOneDot}

\def\wrt{w.r.t\bmvaOneDot}

\begin{document}

\maketitle

\vspace{-15pt}
\begin{abstract}
    %Reducing the amount of supervision required by neural networks is crucial in the context of semantic segmentation, where collecting dense pixel-level annotations is costly. 
    %State-of-the-art 
    Semantic segmentation models have two fundamental weaknesses: i) they require large training sets with costly pixel-level annotations, and ii) they have a static output space, constrained to the classes of the training set. %Ideally we want to extend a model at will, adding new classes, using as few annotations as possible and without access to the original dataset. 
    Toward addressing both problems, we {introduce} a new task, \setting\ (\SET). The goal of \SET\ is to extend a pretrained segmentation model with new classes from few annotated images and without access to old training data.
    %\comment{To address \SET\}
    {To overcome the limitations of existing models in \SET}, we propose \oursFull\ (\ours) that couples prototype learning and knowledge distillation. \ours\ exploits prototypes to %, the first approach for \SET
    %\ours\ uses prototype learning 
    initialize the classifiers of new classes, fine-tuning the network to refine its features representation. 
    We design a prototype-based distillation loss on the scores of both old and new class prototypes to avoid overfitting and forgetting, and batch-renormalization to cope with non-\textit{i.i.d.} few-shot data.
     {We create an extensive benchmark for \SET\ % with different number of classes, images, and learning steps, evaluating several few-shot and incremental learning baselines. Results 
     showing that \ours\ outperforms several few-shot and incremental learning methods in all scenarios.} %few-shot and incremental baselines in \SET
     %\fabio{We will release the our benchmark, the baselines and \ours, the \SET\ benchmark and baselines.}couples batch-renormalization
    %Experiments on two datasets varying the number of new classes, images per class, and incremental learning steps show that \ours\ consistently outperforms several baselines in \SET. %surpass multiple incremental and few-shot learning segmentation algorithms.
\end{abstract}

%-------------------------------------------------------------------------
\section{Introduction}

\begin{figure}[t]
    \centering
    \includegraphics[width=0.9\linewidth]{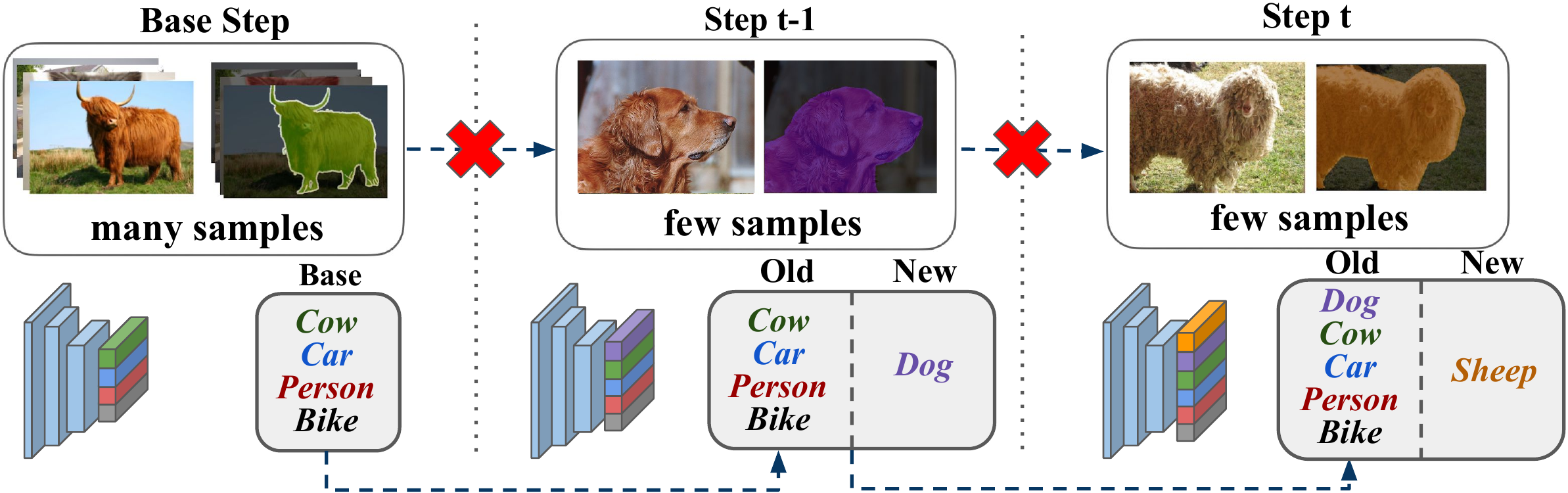}
    \caption{\setting. First, a model is pretrained on a large labeled dataset to learn a set of base classes. Then, in the few-shot learning steps, it learns to segment new classes, given only few annotated images and without access to old datasets.} 
    \label{fig:teaser}
    \vspace{-15pt}
\end{figure}

Deep semantic segmentation models require a large collection of training images with dense pixel-level annotations for classes of interest. However, annotating a large number of images at pixel-level is costly and the output space of the model is restricted to the labeled training classes. Ideally, we want to add new classes to a segmentation model without requiring a large collection of images, but existing methods partly fulfill this aim. Incremental Learning (IL) approaches \cite{michieli2019incremental,cermelli2020modeling} need a large training set to add new classes to a pre-trained model. Few-Shot Semantic Segmentation (FSS) \cite{shaban2017one, rakelly2018few, dong2018few, wang2019panet, zhang2019canet, siam2019adaptive} learns to segment new classes from few images but fully discard old knowledge, while Generalized FSS methods (GFSS) \cite{xian2019spnet} segment both old and new classes, but require access to training data for old classes. This may not be possible \eg if the model is used in a device with limited storage. 

In this work we study a practical scenario where the goal is to learn a segmentation model for both old and new classes with few samples and without access to past training data.
Inspired by object detection \cite{perez2020incremental} and classification \cite{gidaris2018dynamic} literature, we name this problem \textit{\setting} (\SET). This new setting captures different challenges such as learning from few images (as in FSS) to recognize both base and new concepts (as in GFSS) without forgetting old knowledge (as in IL). Fig.~\ref{fig:teaser} illustrates \SET.

To study \SET\ we introduce an evaluation protocol and extensive benchmark on two different datasets, varying the number of classes, images per class, and learning steps. We find that IL and FSS methods struggle on this scenario, either focusing on not forgetting old knowledge \cite{michieli2019incremental, cermelli2020modeling} or failing to adapt the representation on the new classes~\cite{gidaris2018dynamic, qi2018low, siam2019adaptive}. %leading to sub-optimal results due to the poor features extracted for new class pixels}. %

To improve the new class representations while avoiding both forgetting and overfitting, we propose \oursFullAcro\ (\ours), that combines for the first time prototype learning \cite{gidaris2018dynamic, qi2018low} with knowledge distillation \cite{hinton2015distilling}. 
\ours\ exploits prototypes to easily integrate new classes from few-shots, imprinting their pixel-level features as weights on the classifier. Differently form previous few-shot methods \cite{gidaris2018dynamic,qi2018low}, during the few-shot learning (FSL) steps we fine-tune the network end-to-end to improve the feature representation for new class pixels. We prevent both overfitting and forgetting with a novel prototype-based distillation loss that integrates new class scores in the objective. Finally, we find that batch normalization \cite{ioffe2015batch} hurts the performance in \SET\ %during fine-tuning 
since few-shot data are non-\textit{i.i.d.}. We solve this issue by using batch-renorm in the FSL steps \cite{ioffe2017batch}. Experiments demonstrate that \ours\ consistently outperforms all baselines in all \SET\ settings.

\myparagraph{Contributions.} Our contributions are as follows.
(1) We define the \setting\ problem, {requiring learning from few images \cite{shaban2017one, rakelly2018few, dong2018few} while avoiding catastrophic forgetting \cite{mccloskey1989catastrophic, li2017learning, cermelli2020modeling}}; %{We are not aware of previous works studying this semantic segmentation setting.}
%(2) We present \ours, a prototype-based model for \SET. The model combines advances in both few-shot and incremental learning at two levels i) by imprinting classifier weights for new classes, which is fundamental to bootstrap end-to-end training in the incremental learning steps, and ii) with a novel distillation loss that incorporates new class scores in the objective, reducing forgetting while preventing overfitting. 
(2) We present \ours, that overcomes the shortcoming of IL and FSL methods on \SET\ by combining prototype learning (to bootstrap end-to-end training in the FSL steps), knowledge distillation (incorporating new class scores to reduce forgetting while preventing overfitting), and batch-renorm (to cope with non-\textit{i.i.d.} few-shot data). %While prototype learning imprinting classifier weights for new classes, which is fundamental to bootstrap end-to-end training in the few-shot learning steps, and ii) using a novel distillation loss that incorporates new class scores in the objective, reducing forgetting while preventing overfitting.}
(3) We design an extensive benchmark for \SET\ %considering two datasets, and 
%varying the number of classes, images per class, and learning steps, 
and we %, under the same protocol
show that \ours\ consistently outperforms several IL and FSL methods on it. % in all settings. 
The code can be found at \href{http://github.com/fcdl94/FSS}{github.com/fcdl94/FSS}.% to ease research in \SET.}%\SET\ 
%. %: to perform end-to-end training effectively in the few-shot learning steps {thanks to the introduction of a novel knowledge distillation loss}, avoiding both overfitting on new classes and forgetting old knowledge.}

%(3) We benchmark \ours\ on two datasets, varying the number of classes, images per class, and incremental learning step,
%showing that \ours\ 
% consistently outperforming 
% multiple incremental and few-shot learning segmentation algorithms in all %\SET\ 
% scenarios.

%and multiple incremental and few-shot learning segmentation algorithms on \SET\

\section{Related Works}

\vspace{-4pt}\myparagraph{Benchmarks.}
 Similarly to \SET, Few-Shot Segmentation (FSS) \cite{shaban2017one, rakelly2018few, dong2018few, zhang2019canet, wang2019panet, siam2019adaptive, zhang2020sg} aims to segment new classes, given few images depicting them. However, FSS considers an episodic setup \cite{vinyals2016matching}, where the goal is to segment \textit{only} the new classes, often reducing the problem to a binary \cite{shaban2017one, rakelly2018few, siam2019adaptive, zhang2019canet} or 2-way \cite{dong2018few, wang2019panet, zhang2020sg} segmentation one, which is unrealistic. To overcome these limitations, \cite{xian2019spnet} proposed Generalized Few-Shot Segmentation, where the goal is to segment both old and new classes, learning them from several and few images respectively. However, \cite{xian2019spnet} considers an offline scenario, assuming there is always access to all the images. %a large dataset with annotations for old classes.
In contrast, Incremental Learning in segmentation \cite{michieli2019incremental,cermelli2020modeling,li2017learning, siam2019adaptive} assumes to have a large dataset for new classes without access to old datasets. 
\SET\ relies on the intersection of these settings, requiring to learn new classes from a small dataset without accessing old data. Note that settings similar to \SET\ exist in image classification \cite{gidaris2018dynamic, tao2020few, ren2019incremental}, object detection \cite{perez2020incremental} but we are the first to study this setting in semantic segmentation. 
Differences between \SET\ and existing settings are summarized in \tabref{tab:setting-comparison}.
\rev{Concurrently to us, \cite{ganea2021incremental} proposed the incremental few-shot instance segmentation setting. We note that, while being related, instance and semantic segmentation address different challenges, requiring different network architectures and benchmarks.} 

\myparagraph{Semantic Segmentation.}
State-of-the-art % semantic segmentation 
models use a fully convolutional encoder-decoder networks \cite{long2015fully, badrinarayanan2017segnet}, integrate contextual information on pixel-level features in different ways, e.g. through pyramids \cite{zhao2017pyramid,lin2017refinenet,chen2017rethinking, chen2017deeplab, zhang2018exfuse, chen2018encoder}, or attention \cite{yu2020context, yuan2019object, yu2018learning, yuan2018ocnet, fu2019dual, zhang2018context}.
Despite their effectiveness, these models need a large dataset for training, which is often expensive to collect, and they only consider an offline setting, with a static output space. % Here we address both issues, extending the output space of a model using just a few annotated images.

{
\myparagraph{Few-shot Learning} approaches can be split in two groups: {optimization-based} \cite{ravi2016optimization, finn2017model, nichol2018first, rusu2018meta} and metric-learning \cite{snell2017prototypical, gidaris2018dynamic, vinyals2016matching, sung2018learning, chen2017rethinking, qi2018low, wang2019panet, dong2018few, siam2019adaptive}. \ours\ is related to the latter, learning an embedding space where %learn a r
instances of the same class are close to each other. In this context, \cite{snell2017prototypical,gidaris2018dynamic} learned to extract per-class prototypes from few-images through meta-learning. \cite{qi2018low} proposed weight imprinting to add new class weights to a cosine classifier. %used a cosine classifier to represent prototypes,  %propose weight-imprinting by adding novel class embeddings into an existing weight matrix, allowing incremental addition of classes without training.
%\cite{gidaris2018dynamic} %used cosine similarity-based classifiers, 
%learns to generate class weights for new classes through meta-learning. 
\cite{chen2017rethinking} fixed the feature extractor and trained the classifiers for new classes. %, assuming robustness of the frozen feature extractor. 
\cite{dong2018few} extended \cite{snell2017prototypical} on the segmentation task by aggregating pixel-level feature representations. % while \cite{wang2019panet} added an \massi{alignment constraint} on the prototypes. 
\cite{siam2019adaptive} proposed to update also the old classes while computing the new class prototypes. Inspired by them, \ours\ uses prototypes to initialize the classifiers for new classes but, differently, it fine-tunes the whole network using a distillation loss %\fabio{REMOVED BR} %and batch-renorm 
to reduce overfitting and forgetting. % \massi{we can be consistent everywhere, i.e. batch-renorm vs batch renormalization}

\myparagraph{Incremental Learning}
aims to expand the knowledge of a model without forgetting \cite{mccloskey1989catastrophic}.
This problem has been extensively studied in image classification \cite{li2017learning, kirkpatrick2017overcoming,chaudhry2018riemannian,rebuffi2017icarl,hou2019learning,yu2020semantic, douillard2020podnet} and recently in %semantic 
segmentation \cite{ozdemir2019extending,michieli2019incremental,michieli2021knowledge, siam2019adaptive, cermelli2020modeling, douillard2021plop, michieli2021continual}. \cite{michieli2019incremental, michieli2021knowledge} used knowledge distillation \cite{hinton2015distilling}, to enforce output consistency between the current model %the model to output features consistent 
and the one at the previous learning step. %\massi{add what previous approaches did, e.g. distillation stuff} 
\cite{cermelli2020modeling} investigated the background shift, revisiting classification and distillation terms. While also \ours\ employs a distillation loss, we couple it with prototype learning to i) effectively initialize the classifier for new classes; ii) avoid overfitting on few images. 
%\fabio{Maybe we can discuss that IL methods deal with background shift, while we don't}
% \fabio{We may also cite Lomonaco for BR}
}

\begin{table}[t]
\centering
    \setlength{\tabcolsep}{20pt} % Default value: 6pt
    \resizebox{\linewidth}{!}{
\begin{tabular}{l|c c c c c}
                        \multirow{2}{*}{Semantic Segmentation}&       \multicolumn{2}{c}{Training}    &   \multicolumn{2}{c}{Output}  \\ 
                        & data & few-shot& class& multi-step\\\hline
                        Offline &$\set D^0$& -&$\set C^0$& - \\
                         \hline
Few-Shot     \cite{shaban2017one, rakelly2018few, dong2018few, zhang2019canet}         & $\set D^{t}$  &    \cmark &$\set K^t$  & - 
\\
Generalized Few-Shot  \cite{xian2019spnet}& $\cup_{s=0}^{t}\set D^{s}$  & \cmark& $\set C^t$& - \\
Incremental Learning \cite{michieli2019incremental,cermelli2020modeling}& $\set D^t$&-&$\set C^t$&\cmark\\\hline
\textbf{Incremental Few-Shot}     & $\set D^t$  &\cmark&  $\set C^t$   & \cmark \\
\end{tabular}}
\vspace{-5pt}
\caption{Comparing different semantic segmentation settings. $t$ denotes the current learning step, $\set K^t$ denotes all classes labeled in the dataset $\set D^t$ while $\set C^t=\cup_{s=0}^{t}\set K^{s}$.} 

\label{tab:setting-comparison}
\vspace{-15pt}
\end{table}

\section{\setting\ (iFSS)} \label{sec:setting}
%In this section, we formalize the \setting\ problem (\SET), and discuss its differences with existing settings, %\ie (Generalized) Few-shot Segmentation and Incremental Learning, 
%summarized in \tabref{tab:setting-comparison}.

%\myparagraph{Problem Definition.} 
The goal of \SET\ is to learn a model that assigns to each pixel of an image its corresponding semantic label in a set $\set C$. Differently from standard semantic segmentation, $\set C$ is expanded over time using few images with pixel-level annotations of new classes. 

Formally, let us denote as $\set C^t$ the set of semantic categories known by the model after learning step $t$, where \textit{learning step} denotes a single update of the model's output space. During training we receive a sequence of datasets $\{\set D^0, \dots, \set D^T\}$ where $\set D^t = \{(x, y)| x \in \set X\,, y \in \set Y^t\}$. In each $\set D^t$, $x$ is an image in the space $\set X\in \real^{|I|\times 3}$, with $I$ the set of pixels, and $y$ its corresponding label mask in $\set Y^t \subset \set (C^t)^{|I|}$. Note that $\set D^0$ is a large dataset while $\set D^t$ are few-shot ones, \ie $|\set D^0|\gg|\set D^t|$, $\forall t\geq1$. The model is first trained on the large dataset $\set D^0$ and % containing pixel level annotations for several semantic categories. Then, i
incrementally updated with few-shot datasets. % $\set D^t$ is available. 
We name the first learning step on $\set D^0$ as the \textit{base} step. 
Note that at step $t$ the model has access only to $\set D^t$. %, with $\set C^0$ as the base classes. 
 % This implies that at learning step $t+1$ we have access only to the dataset $\set D^{t+1}$.

From the formulation, we make two assumptions: i) each dataset contains annotations for new classes, \ie $\set C^t \subset \set C^{t+1}$; ii) pixels of old classes $\set C^t$ will be labeled in $\set D^{t+1}$, but \textit{only} if present. %, since annotating few images is cheap. 
{Note that fully annotating few images is cheap, in contrast to \cite{cermelli2020modeling} where $\set D^{t+1}$ is large and annotating pixels of both old and new classes is expensive. } %only pixels of new classes are annotated due to the dataset $\set D^{t+1}$ is large }.

\begin{figure}
    \centering
    \includegraphics[width=\linewidth]{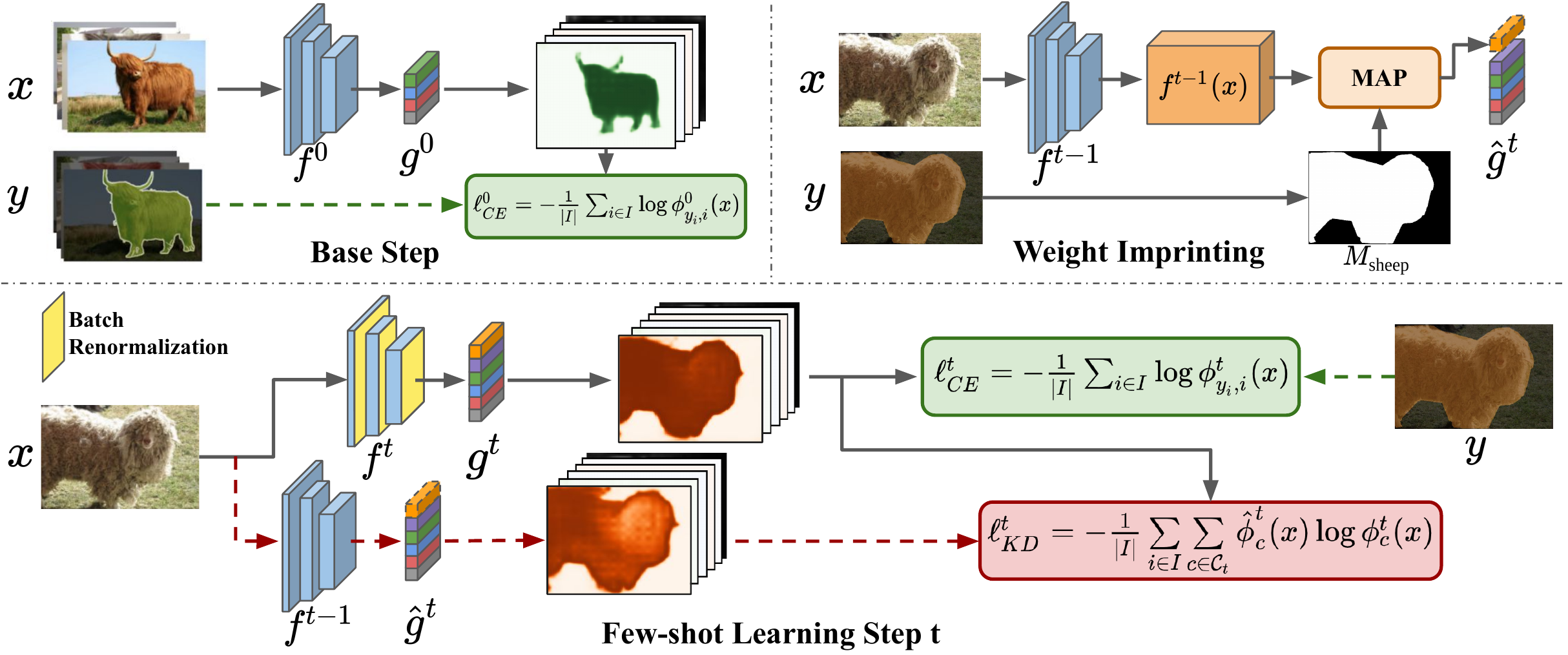}\vspace{-5pt}
    \caption{Illustration of \ours. {In the base step (top left), we train a prototype-based model using the cross-entropy loss $l_{CE}$.} Then (top right), given few images of a new  class, we initialize its prototype with %update the classifier using 
    Masked Average Pooling (MAP). We then fine-tune the network using the CE loss and our prototype-based knowledge-distillation ($l_{KD}$) to preserve old knowledge while reducing overfitting the new class representation (bottom). To cope with the non-\textit{i.i.d.} few-shot data, we use batch-renorm in place of batch-norm in the few-shot learning steps.} 
    \label{fig:method}
    \vspace{-10pt}
\end{figure}

\section{Prototype-based for iFSS} \label{sec:method}
In this section, we present \textit{\oursFull} (\ours), illustrated in \figref{fig:method}.
% \ours\ uses prototype learning to %learn class prototypes and 
% effectively integrate new classes from few-shots by imprinting their prototypes in the classifier.
% In the few-shot learning (FSL) step, the model is trained end-to-end to improve the network representation, reducing both forgetting and overfitting exploiting a novel prototype-based distillation loss.
In the base step, \ours\ learns a prototype-based model using a standard training procedure. 
In the few-shot learning (FSL) steps, it first exploits prototype learning to initialize the classifiers weights 
for new classes, and then fine-tunes the network end-to-end with a
prototype-based distillation loss, % reducing both forgetting and overfitting, 
using batch-renorm to cope with non-\textit{i.i.d.} data.
%Figure \ref{fig:method} gives an overview of the approach. 

\subsection{Prototype Learning}
\label{sec:proto-init} 
\vspace{-4pt}\myparagraph{Learning a prototype-based model.}
Our goal is to learn a model $\phi^t$ that maps each pixel to a probability distribution over the set of classes, \ie $\phi^t : \set X \to \real^{|I|\times |\set C^t|}$, where $t$ denotes the last learning step. 
We assume $\phi^t$ composed of a feature extractor $f^t: \set X \to \real^{|I|\times d}$ and a classifier $g^t: \real^{|I|\times d}  \to \real^{|I|\times |\set C^t|}$, such that $\phi^t = g^t \circ f^t$. Here, $d$ is the feature dimension and $g^t$ is a softmax classifier with parameters $W^t = [w^t_1, \dots, w^t_{|\set C^t|}] \in \real^{d \times |\set C^t|}$. %Note that state-of-the-art segmentation architectures \cite{chen2017deeplab, zhao2017pyramid} fit in this definition. 

In the base step, we want to prepare $\phi^0$ to include new classes given few examples. %To this end, we need our model to i) extract as much information as possible from the few available images while ii) avoiding biases toward either old or new classes. 
To this aim, we enforce the classifier weights to represent class prototypes. The prototypes shall reflect the average pixel-level features of a class, so that the features extracted from (few) pixels of the new classes provide a good estimate of %used to directly initialize 
their corresponding classifier weights. Following previous works \cite{gidaris2018dynamic, qi2018low}, we achieve this through a cosine classifier. 

Formally, in the base %learning 
step, we train the network with a cross-entropy loss over all pixels: %the pixel-level class scores: %cosine similarities among each feature vector and the class prototypes:
\begin{equation}
\label{eq:objective}
    \ell^0_{CE}(x,y) = -\frac{1}{|I|} \sum_{i \in I} \log \phi_{y_i,i}^0(x)
\end{equation}
where $\phi_{c,i}^0(x)$ is the probability of pixel $i$ of $x$ to belong to class $c$. The score $\phi_{c,i}^t(x)$ % for a class $c$ in pixel $i$ 
is computed as a softmaxed cosine similarity between the features and the class prototype $w^t_c$:
\begin{equation}
    \phi_{c,i}^t(x) = g_{c,i}^t(f^t(x)) = \frac{e^{s_{c,i}^{t}}}{\sum_{k \in \set C_0} e^{s_{k,i}^{t}}}, \;\;\;    s_{c,i}^{t} = \tau \frac{f_i^t(x)^\intercal  w^t_c}{||f_i^t(x)||\ ||w^t_c||}
\end{equation}
where %$w^t_c$ being the classifier weight for class $c$, 
$f_i^t(x)$ are the features extracted at pixel $i$, and $\tau$ is a scalar which scales the similarity in the range $[-\tau, \tau]$. 
With the cross-entropy loss in Eq.~\eqref{eq:objective}, the model minimizes the cosine distance between a prototype and the features of %a given 
its class, ensuring their % and the prototype of the ground-truth class. As a consequence, the classifier capturing the 
compatibility. % between feature vectors and class prototypes.

\myparagraph{Initializing prototypes of new classes.}
% We need to first initialize the classifier's parameters for each new class in $K^t$. Since the objective in \equref{eq:objective} ensures that the classifier weights represent class prototypes, we can 
Given a dataset $\set D^t$, let us denote as $\set K^t$ the set of new classes, \ie $\set K^t = \set C^t \setminus \set C^{t-1}$. After the base step, features of a class $k\in\set K^t$ provide an estimate of the prototype $w_k$. Thus, we compute the new class prototypes by aggregating the %set of 
features extracted for each pixel of the class $k$ present in images of $\set D^t$. %\massi{, performing weight imprinting \cite{qi2018low}}. %. After the base step, we initialize the parameters for each $k\in\set K^t$ as their %respective 
%prototypes, through \textit{weight imprinting} \cite{qi2018low}. 
%Since features of a class $k$ provide an estimate of the prototype $w_k$, we compute the new class prototypes by aggregating the %set of 
%features extracted for each pixel of the class $k$ in $\set D^t$. % corresponding to each pixel $i$ of a class $k$ in each new image $x \in \set D^t$. 
Inspired by \cite{siam2019adaptive, dong2018few, wang2019panet}, we use masked average pooling (MAP) to initialize the prototypes:
%MAP consists in computing the  average of all the feature vectors in $\set F_k^t$ corresponding to pixels of the class of interest $k$. Given the set $\set D^t$, 
%We compute MAP for class $k$ as:
\begin{equation}\label{eq:map}  w^t_k = 
    \text{MAP}_k({\set D^t}) = \frac{1}{|\set D^{t}_k|}\sum_{(x,y) \in \set D^{t}_k}\frac{\sum_{i \in I} M_{k,i}(y) \frac{f^t_i(x)}{||f^t_i(x)||}}{\sum_{i \in I} M_{k,i}(y)}, 
\end{equation}
where $M_{k}(y)$ is a binary mask indicating which pixels belong to class $k$, and $\set D^{t}_k$ is the set of images in $\set D^t$ containing at least one pixel of class $k$. As we will show experimentally, this strategy provides a good initial estimate of the classifier that is not needed in standard IL but is crucial in \SET\ for learning to segment new classes from few samples. 
%The MAP can be directly employed as the classifier weights of each new class $k \in \set K^t$, \ie $w^t_k = \text{MAP}_k({\set D^t})$.

\subsection{Distilling Prototypes for \SET}
%As we will show experimentally, and witnessed by previous works \cite{qi2018low, gidaris2018dynamic}, WI is a general strategy for few-shot learning, whose effectiveness is independent on the particular setting and task. However, 
Although %weight imprinting is a good classifier initialization strategy, 
prototypes are good to initialize the classifier, relying on a feature extractor tuned on different semantic classes (\ie $\set C^{t-1}$) is suboptimal. %, due to the limited since the ability of the network to extract good features for unseen classes is limited. 
To refine the feature representation, we train the model \textit{end-to-end} in the few-shot learning steps. As pointed out in \cite{gidaris2018dynamic,perez2020incremental}, end-to-end training in FSL steps may lead to overfitting and forgetting. %\cite{gidaris2018dynamic} and forgetting %old knowledge
%\cite{perez2020incremental}. %However, while na\"ive end-to-end training would produce poor results,
We address these issues %borrowing ideas from the IL literature. In particular, we propose to use  
by designing a distillation loss on the prototypes that regularizes the training, reducing both overfitting and catastrophic forgetting. We also use batch-renorm~\cite{ioffe2017batch} to cope with the non \textit{i.i.d}. few-shot data. In the following we describe the two components. %The ideal solution would be to fine-tune the network on new class data. However, 
% previous works \cite{gidaris2018dynamic, perez2020incremental} point out that fine-tuning should not be performed on incremental few-shot  
% since it is prone to (i) over-fitting the new classes, due to the limited amount of data \cite{gidaris2018dynamic}
% , and (ii) catastrophic forgetting \cite{mccloskey1989catastrophic}, since without revisiting the data for previous knowledge the network is biased toward new data \cite{perez2020incremental}. Here we take a different perspective and we argue that end-to-end training on few-shot network can largely boosts the performance of the model. This is possible if i) we keep into account the incremental nature of the task, borrowing principles of IL to avoid catastrophic forgetting and ii) we devise a strategy to address the shift in the data distributions caused by the non i.i.d. few-shot samples. 

\myparagraph{Prototype-based Distillation.} %We start by describing how we design our end-to-end training objective for every few-shot learning step $t$, with $t\geq1$. In this scenario, 
Given a pair $(x,y)\in \set D^t$, we update $\phi^t$ by minimizing: % the following loss function:
\begin{equation}
\label{eq:objective-few-shot}
    \ell^t(x,y) = \ell^t_{CE}(x,y) + \lambda \ell^t_{KD}(x, \phi^t,\Phi)
\end{equation}
where $\lambda$ is a %trade-off
hyperparameter, $\ell^t_{CE}$ is the %cross-entropy 
loss of Eq.~\eqref{eq:objective} over $\set C^t$ and $\ell^t_{KD}$ is a knowledge distillation loss, where $\Phi$ is the teacher model. While previous works \cite{michieli2019incremental, cermelli2020modeling} directly use a copy of the network after the previous learning step as teacher, \ie $\Phi = \phi^{t-1}$, here we exploit the benefits of prototype learning, defining $\Phi$ as the network after the initialization of the new class prototypes. In particular, we set $\Phi = \hat{\phi}^t$, where $\hat{\phi}^t= \hat{g}^t \circ f^{t-1}$ and the parameters $\hat{W}^t = [\hat{w}^t_1, \dots, \hat{w}^t_{|\set C^t|}]$ of $\hat{g}^{t}$ as:
%$\phi^{t-1}=g^{t-1}\circ f^{t-1}$ is a copy of the network after the previous learning step.
% In Eq.~\eqref{eq:kd-general}, %the teacher model 
% the teacher $\Phi$ should force %has the role of forcing 
% $\phi^t$ to extract representations consistent with the old knowledge, %that do not deviate from the discriminative ones, 
% reducing catastrophic forgetting.
% Previous works \cite{michieli2019incremental, cermelli2020modeling} directly use a copy of the network after the previous learning step as teacher, \ie $\Phi = \phi^{t-1}$, %=g^{t-1}\circ f^{t-1}$, 
% instantiating $d$ as L2 at the feature-level \cite{michieli2019incremental}, or cross-entropy on the probabilities of old classes \cite{li2017learning,cermelli2020modeling}. 
% Differently, we exploit the benefits of prototype learning, defining $\Phi$ as the network after the initialization of the prototypes $\hat{W}^t$ of $\hat{g}^{t}$ for new class $k\in \set K^t$ (see \equref{eq:map}), such that $\Phi = \hat{\phi}^t = \hat{g}^t \circ f^{t-1}$, and $d$ as
\begin{equation}
\hat{w}_k^t = 
\begin{cases}
w^{t-1}_k,   &\text{if} \;k \in \set C^{t-1}\\
\text{MAP}_k({\set D^t}) & \text{otherwise.}
\end{cases}
\end{equation}
We then define the distillation loss $\ell^t_{KD}$ as: %The old model $\phi^{t-1}$ acts as a teacher and the current model $\phi^t$ as a student, minimizing the objective:
\begin{equation}
    \label{eq:self-distill}
    % \ell^t_{KD}(x, \phi^t, \Phi)  = \frac{1}{|I|} \sum_{i \in I} d(\phi^{t}(x_i), \Phi(x))
    \ell^t_{KD}(x, \phi^t, \Phi)  = -\frac{1}{|I|} \sum_{i \in I} \sum_{c\in\set C^t} \Phi_c^t(x)\log \phi_c^t(x).
\end{equation}
% \comment{In Eq.~\eqref{eq:kd-general}, the old model $\phi^{t-1}$ acts as a teacher and the current model $\phi^t$ as a student, forcing $\phi^t$ to extract representations that do not deviate from the discriminative ones for the old knowledge in $\phi^{t-1}$, reducing catastrophic forgetting. While previous works in IL %for semantic segmentation 
% instantiate $d$ as either L2 at the feature-level \cite{michieli2019incremental}, or cross-entropy on the probabilities of old classes \cite{cermelli2020modeling}, here we fully exploit the good initialization given by the prototype-based classifier.}

 %{and our distance function as:}
% \begin{equation}\label{eq:self-distill}
%     d(\phi^t(x), \hat{\phi}^t(x)) = -\sum_{c\in\set C^t} \hat{\phi}_c^t(x)\log \phi_c^t(x).
% \end{equation} 
Note that in Eq.~\eqref{eq:self-distill}, we explicitly consider the scores that the teacher produces for both old classes in $\set C^{t-1}$ and \textit{new} ones in $\set K^t$. The advantage of this new formulation \wrt standard knowledge distillation in IL is that we not only alleviate forgetting by forcing the current model to keep scores for old classes similar to the old model, but we also encourage the prototypes of new classes to be close to their initial estimate given by $f^{t-1}$. This allows the model to reduce overfitting on the few-shot data of new classes, a main problem in \SET. 
% This forces the model to keep scores for old classes similar to the old model, alleviating forgetting, and encourages the prototypes of new classes to be close to their initial estimate given by $f^{t-1}$, reducing overfitting. %In fact, forces $\phi^t$ to not deviate from the old feature extractor (since prototypes are initialized from $f^{t-1}$) while reducing overfitting on local information of the few-shot dataset by forcing coherence with the . %. The latter follows the principle of \textit{refining} the model for new classes % where the latter is obtained through the classifier initialized with weight imprinting. 
% The advantage of this new formulation \wrt standard knowledge distillation in IL \cite{michieli2019incremental,cermelli2020modeling,rebuffi2017icarl,li2017learning} is that we address forgetting while also reducing overfitting on the few-shot data of new classes, a main problem in \SET. 

\myparagraph{Coping with non-i.i.d. data.}
Despite the regularized training, we found that in extreme few-shot scenarios (\eg 1-shot settings) a main cause of the drop in performance is the drift of statistics in the batch-normalization (BN) \cite{ioffe2015batch} layers of the network. BN assumes independent and identically distributed (\textit{i.i.d.}) data \cite{ioffe2015batch, ioffe2017batch} but in the FSL steps we have small datasets where most pixels belong to new classes, thus the input is inherently non \textit{i.i.d.}. Updating the statistics on this non-\textit{i.i.d.} set makes them poor and biased.

Two simple solutions are either using %to normalize the FSL data either with 
the global BN statistics of the base step, % on $\set D_0$, 
or the training batch ones but without updating their global estimate used at test time. However, we found the first solution causing training instability and the second poor performance due to misalignment between features extracted for the new classes at training and test time. %without normalizing the features during training we found that the gradients explode, making end-to-end training unfeasible. Another solution may be to freeze the BN statistics during inference but not during the FSL step. However, this solution does not bring improvements since the network will be trained using the (non \textit{i.i.d.}) batch statistics, which differ from the ones collected in the initial training stage, thus introducing a misalignment between features extracted for the new classes between training and test time. 

Ideally, we want to normalize features in the FSL step while i) avoiding the shift of the statistics toward the new class data and  ii) aligning training and inference statistics. % with the inference ones. %, as extracted by the main architecture. 
To achieve this, we take inspiration from continual learning works with non-\textit{i.i.d.} data \cite{lomonaco2020rehearsal} and use
batch-renorm (BR) \cite{ioffe2017batch}.
% In practice, BN layers normalize each value of the input $z^c_i$, where $c$ indicates the channel and $i$ its spatial position, to have a learnable mean and standard deviation:
% \begin{equation}
%     \hat{z}^c_i = \gamma \frac{z^c_i - \mu^c}{\sigma^c} + \beta,
% \end{equation}
% where $\gamma$ and $\beta$ are the learnable parameters, and $\mu^c$ and $\sigma^c$ are, respectively, the mean and standard deviation across all the images in the batch on the channel $c$. 
% During inference, the BN layers replace the batch statistics with the running statistics, which are the moving averages of the mean $\mu^c_r$ and standard deviation $\sigma^c_r$:
% \begin{equation}
%     \hat{z}^c_i = \gamma \frac{z^c_i - \mu^c_r}{\sigma^c_r} + \beta.
% \end{equation}
Batch-renorm (BR) revisits BN by normalizing a feature $q$ with the running statistics in place of the training batch statistics:
\begin{equation}
    \hat{z}^q_i = \gamma (\frac{z^q_i - \mu^q}{\sigma^q} \frac{\sigma^q}{\sigma_r^q} + \frac{\mu^q - \mu^q_r}{\sigma_r^q}) + \beta,
\end{equation}
where $\gamma$ and $\beta$ are learnable parameters, $\mu_r^q$ and $\mu^q$ the global and batch mean, and $\sigma_r^q$ and $\sigma^q$ the global and batch standard deviation. %and global one and $\sigma^c$, are, respectively, the mean and standard deviation across all the images in the batch on the channel $c$. It is important to note that the gradient does not flow in $r$ and $d$, so the normalization effect on the gradient is still active. 
We freeze $\mu_r^q$ and $\sigma_r^q$ after the base step to prevent them from shifting toward new classes and damaging the performance on base ones. %\massi{BR, shall we cite?}%As we will show experimentally, this simple change is crucial to perform stable end-to-end training in the FSL step.  

\section{\SET\ Experiments} \label{sec:experiments}
%In this section, we compare \ours\ with multiple baselines adapted from related scenarios in two scenarios: single step, \ie all the new classes are added together, and sequential addition of new classes, \ie one-by-one.
%In this section, we first define the experimental protocol and we then compare \ours\ with multiple few-shot and incremental learning methods.

\vspace{-4pt}\myparagraph{Experimental Protocol.}
To assess \SET\ performance of a model, we need (i) a large dataset containing an initial set of classes and (ii) one or more few-shot datasets containing new classes. % which represent in a heterogeneous way the main challenges of the new benchmark.
We create such experimental setting on %two datasets, 
Pascal-VOC 2012 (VOC) \cite{pascal-voc-2012} containing 20 classes, and COCO \cite{caesar2018coco, lin2014microsoft} where, as in FSS works \cite{zhang2019canet, wang2019panet}, we use the 80 thing classes. % following FSS literature . %we use only the 80 thing classes.
{Following FSS works \cite{shaban2017one, nguyen2019feature, zhang2019canet}, we consider 15 and 60 of the classes as \textit{Base} ($\set C^0$) and 5 and 20 as \textit{New} ($\set C^t \backslash \set C^0$), for VOC and COCO respectively}. %of the classes as \textit{Base} ($\set C^0$) and $\frac{1}{4}$ as \textit{New} ($\set C^t \backslash \set C^0$), and 
We propose two protocols, each starting with pretraining on \textit{Base} classes: in one there is a single FSL step on all \textit{New} classes, while in the other we have multiple %learning 
steps: 5 steps of 1 class on VOC and 4 steps of 5 classes on COCO.
%We consider, respectively for VOC and COCO, 15 and 60 of the classes as \textit{Base} ($\set C^0$) and 5 and 20 as \textit{New} ($\set C^t \backslash \set C^0$). We propose two protocols, each starting with pretraining on \textit{Base} classes: in one there is a single FSL step on all \textit{New} classes, while in the other we have multiple learning steps, 5 steps of 1 class on VOC and 4 steps of 5 classes on COCO.
%Moreover, 
We divide VOC in 4 folds of 5 classes and COCO in 4 folds of 20 classes, running experiments 4 times by considering each fold in turn as the set of new classes. \rev{We report the list of classes of each fold in the supplementary material.} We name the single-step settings VOC-SS and COCO-SS, and the multi-step VOC-MS and COCO-MS.

For each setting, we consider 1, 2 or 5 images in the FSL step and we average the results of multiple trials, each using a different set of images. \rev{The images are randomly sampled from the set of images containing at least one pixel of the new class, {without} imposing any constraint about the presence of old classes. The images contain annotations for the new class they are sampled for and, if present, also for previous classes.}
{Since we do not follow an episodic setup, during the FSL step we only rely on the provided few-shot images (both for weight-imprinting and for training) without using other images.}
To ensure that the model does not use pixels from new classes in the base step, we exclude from the initial dataset all the images containing pixels of new classes. 
Finally, we report the results on the whole validation set of each dataset, considering all the seen classes.
%Since different images in the FSL step may lead to different results, we average the results of multiple trials, each with a different set of images.

Following the protocol of \cite{xian2019spnet} for GFSS, we assess a method's performance using three metrics based on the mean intersection-over-union (mIoU) \cite{pascal-voc-2012}: %, \ie the standard metric of semantic segmentation: 
mIoU on base classes (\textit{mIoU-B}), mIoU on new classes (\textit{mIoU-N}), and the harmonic mean of the two % on base and new classes mIoU 
(\textit{HM}). %, where we refer to $\set C^0$ as \textit{Base} classes, and to $\set C^t \backslash \set C^0$ as the \textit{New} ones. 
{As in \cite{cermelli2020modeling, michieli2019incremental}, we always report the results after the last FSL step.}
For space reasons, we report the average performance on all folds here and the results on each fold % separately
in the supplementary.
%For the sake of space, here we report the average performance on all folds and we report the results for each fold in the supplementary material.

\myparagraph{Baselines.}
% took inspiration from https://openaccess.thecvf.com/content_CVPR_2020/papers/Perez-Rua_Incremental_Few-Shot_Object_Detection_CVPR_2020_paper.pdf
We consider %compare \ours\ %on the \SET\ setting 
9 baselines in our benchmark: three few-shot classification (FSC) methods \cite{qi2018low, gidaris2018dynamic, tian2020rethinking}, two (G)FSS methods \cite{siam2019adaptive,xian2019spnet}, three IL methods \cite{li2017learning, michieli2019incremental, cermelli2020modeling} and na\"ive fine-tuning (FT).  These models are %We choose them because they are 
either state-of-the-art in their respective settings \cite{gidaris2018dynamic, tian2020rethinking,siam2019adaptive,xian2019spnet,michieli2019incremental, cermelli2020modeling} or simple yet effective baselines (e.g.\cite{qi2018low,li2017learning}). % either proposed or adapted from related settings. 
The adapted FSC methods are %for \SET, \cite{qi2018low, gidaris2018dynamic, tian2020rethinking}. 
Weight-imprinting \cite{qi2018low} (WI, Sec.~\ref{sec:proto-init}); Dynamic WI \cite{gidaris2018dynamic} (DWI), an attention-based variant of WI; Rethinking FSL \cite{tian2020rethinking} (RT), that %that trains the network on the base step and then 
fine-tunes only the classifier weights for new classes. %

From FSS, we compare with Adaptive Masked Proxies \cite{siam2019adaptive} (AMP), a variant of WI %for FSS 
that updates also classifier weights of old classes, and Semantic Projection Network \cite{xian2019spnet} (SPN) %. AMP is  %taken into account to refine the classifier weights for new classes. %computes the weight for the classifier given the features of new classes as in WI, but it computes also the weight for seen classes, updating the classification weight. 
%SPN \cite{xian2019spnet} is 
 a GFSS method that projects visual features to a semantic space (\ie word embeddings). % for GFSS.

The IL methods are Learning without Forgetting (LwF) \cite{li2017learning}, applying knowledge distillation (KD) \cite{hinton2015distilling} on the old class probabilities; Incremental Learning Techniques (ILT) \cite{michieli2019incremental}, performing KD also at feature-level for segmentation; % directly at feature-level; 
and Modeling the Background (MiB) \cite{cermelli2020modeling} revisiting standard classification and distillation losses to address the background shift. \rev{Note that, when old classes are annotated in FSL steps, the revised cross-entropy of MiB reduces to the standard cross-entropy formulation.}

\myparagraph{Implementation details.}
In all experiments we use the Deeplab-v3 \cite{chen2018encoder} with ResNet-101 \cite{he2016deep}, following %the implementation of 
\cite{rota2018place} to reduce the memory footprint. We use ResNet-101 with ASPP as feature extractor and %consisting of the , while the
%$g$ is 
a 1$\times$1 convolutional layer as classifier.
% To reduce the memory footprint, we use in-place activate batch normalization \cite{rota2018place}, upon which we also implemented batch renormalization.
% In all experiments we use the Deeplab-v3 architecture \cite{chen2018encoder} with ResNet-101 \cite{he2016deep} as the backbone of the encoder and an output stride of 16. We use ResNet-101 with ASPP as feature extractor $f$, %consisting of the , while the
% classifier $g$ is a 1$\times$1 convolutional layer.
% To reduce the memory footprint, we use in-place activate batch normalization \cite{rota2018place}, upon which we also implemented batch renormalization.
As it is standard practice in %semantic segmentation \cite{chen2018encoder, zhao2017pyramid, zhang2018exfuse} and 
FSS and IL \cite{dong2018few, shaban2017one, wang2019panet, xian2019spnet, siam2019adaptive,cermelli2020modeling}, we initialize the ResNet backbone using an ImageNet pretrained model.
All baselines have been re-implemented by us and share the same segmentation network and training protocols to ensure a fair comparison.
We compute the results using single-scale full-resolution images, without any post-processing. {The code will be released.}
% We will publicly release the code with all benchmarks and baselines.

% For training, we adopted the same protocol of \cite{chen2018encoder}, training the network with SGD and the same learning rate policy, momentum, weight decay and data augmentation, both on the base and few-shot steps.
% We train for 30 epochs on Pascal-VOC and 20 epochs on COCO using a learning rate $10^{-2}$ and batch size 24 on the base step. 
% In the FSL step, considering the addition of a batch of classes, we train using batch size ${min}(10, |\set D_n|)$ for 1000 iteration with learning rate $10^{-3}$. 
% Instead, considering the sequential addition of new classes, we train using the batch size ${min}(10, |\set D_n|)$ for 200 iteration every class with learning rate $10^{-4}$.\barbara{Fabio please clarify the last sentence}

\begin{table*}[t]
    \centering
    \setlength{\tabcolsep}{3pt} % Default value: 6pt
    \resizebox{\linewidth}{!}
    {\begin{tabular}{ll|ccc|ccc|ccc||ccc|ccc|ccc}
    & & \multicolumn{9}{c||}{\textbf{VOC-SS}} & \multicolumn{9}{c}{\textbf{COCO-SS}} \\
    &  &  \multicolumn{3}{c}{\textbf{1-shot}} & \multicolumn{3}{c}{\textbf{2-shot}} & \multicolumn{3}{c||}{\textbf{5-shot}} & \multicolumn{3}{c}{\textbf{1-shot}} & \multicolumn{3}{c}{\textbf{2-shot}} & \multicolumn{3}{c}{\textbf{5-shot}} \\ \hline
& Method	& mIoU-B & mIoU-N	 & HM	 & mIoU-B & mIoU-N	 & HM & mIoU-B & mIoU-N	 & HM  & mIoU-B & mIoU-N	 & HM	 & mIoU-B & mIoU-N	 & HM & mIoU-B & mIoU-N	 & HM \\ \hline
& FT                                    & 58.3&	9.7&	16.7&	59.1&	19.7&	29.5&	55.8&	29.6&	38.7 & 41.2&	4.1&	7.5&	41.5&	7.3&	12.4&	41.6&	12.3&	19.0 \\ \hline
\parbox[t]{2mm}{\multirow{3}{*}{\rotatebox[origin=c]{90}{FSC}}}
& WI \cite{qi2018low}                 & 62.7&	15.5&	24.8&	63.3&	19.2&	29.5&	63.3&	21.7&	32.3 & 43.8&	6.9&	11.9&	44.2&	7.9&	13.5&	43.6&	8.7&	14.6 \\
& DWI \cite{gidaris2018dynamic}       & \textbf{64.3}&	15.4&	24.8&	\textbf{64.8}&	19.8&	30.4&	64.9&	23.5&	34.5 & 44.5&	7.5&	12.8&	45.0&	9.4&	15.6&	44.9&	12.1&	19.1 \\
& RT \cite{tian2020rethinking}        & 59.1&	12.1&	20.1&	60.9&	21.6&	31.9&	60.4&	27.5&	37.8 & \textbf{46.2}&	5.8&	10.2&	\textbf{46.7}&	8.8&	14.8&	{46.9}&	13.7&	21.2 \\ \hline
\parbox[t]{2mm}{\multirow{2}{*}{\rotatebox[origin=c]{90}{FSS}}}
& AMP  \cite{siam2019adaptive}        & 57.5&	16.7&	25.8&	54.4&	18.8&	27.9&	51.9&	18.9&	27.7 & 37.5&	7.4&	12.4&	35.7&	8.8&	14.2&	34.6&	11.0&	16.7 \\
& SPN  \cite{xian2019spnet}           & 59.8&	16.3&	25.6&	60.8&	26.3&	36.7&	58.4&	\textbf{33.4}&	42.5 & 43.5&	6.7&	11.7&	43.7&	10.2&	16.5&	43.7&	15.6&	22.9 \\ \hline
\parbox[t]{2mm}{\multirow{3}{*}{\rotatebox[origin=c]{90}{IL}}}
& LwF  \cite{li2017learning}          & 61.5&	10.7&	18.2&	63.6&	18.9&	29.2&	59.7&	30.9&	40.8 & 43.9&	3.8&	7.0&	44.3&	7.1&	12.3&	44.6&	12.9&	20.1 \\
& ILT  \cite{michieli2019incremental} & \textbf{64.3}&	13.6&	22.5&	64.2&	23.1&	34.0&	61.4&	32.0&	42.1 & \textbf{46.2}&	4.4&	8.0&	46.3&	6.5&	11.5&	\textbf{47.0}&	11.0&	17.8 \\
& MiB  \cite{cermelli2020modeling}    & 61.0&	5.2&	9.7&	63.5&	12.7&	21.1&	\textbf{65.0}&	28.1&	39.3 & 43.8&	3.5&	6.5&	44.4&	6.0&	10.6&	44.7&	11.9&	18.8 \\ \hline
%\ours	& 61.2&	18.4&	28.3&	60.2&	26.4&	36.7&	59.3&	34.3&	43.5 & 41.9&	7.9&	13.3&	42.8&	10.8&	17.3&	43.8&	15.5&	22.9 \\
% WIT     & 56.6&	14.0&	22.5&	57.1&	23.5&	33.3&	56.0&	31.6&	40.4 & 39.9&	7.4&	12.5&	40.8&	10.4&	16.5&	42.1&	15.1&	22.3 \\
& \textbf{\ours}                      & 60.9&	\textbf{18.6}&	\textbf{28.4}&	60.5&	\textbf{26.4}&	\textbf{36.8}&	60.0&	\textbf{33.4} &	\textbf{42.8} & 40.8&	\textbf{8.2}&	\textbf{13.7}&	40.9&	\textbf{11.1}&	\textbf{17.5}&	42.8&	\textbf{15.7}&	\textbf{23.0} \\     
    \end{tabular}}
    \vspace{-4pt} \caption{\SET: mIoU on single few-shot learning step scenarios.} \label{tab:results-single}
    \vspace{-15pt}
\end{table*}
\begin{figure*}[t]
    \centering
    \includegraphics[width=0.9\linewidth]{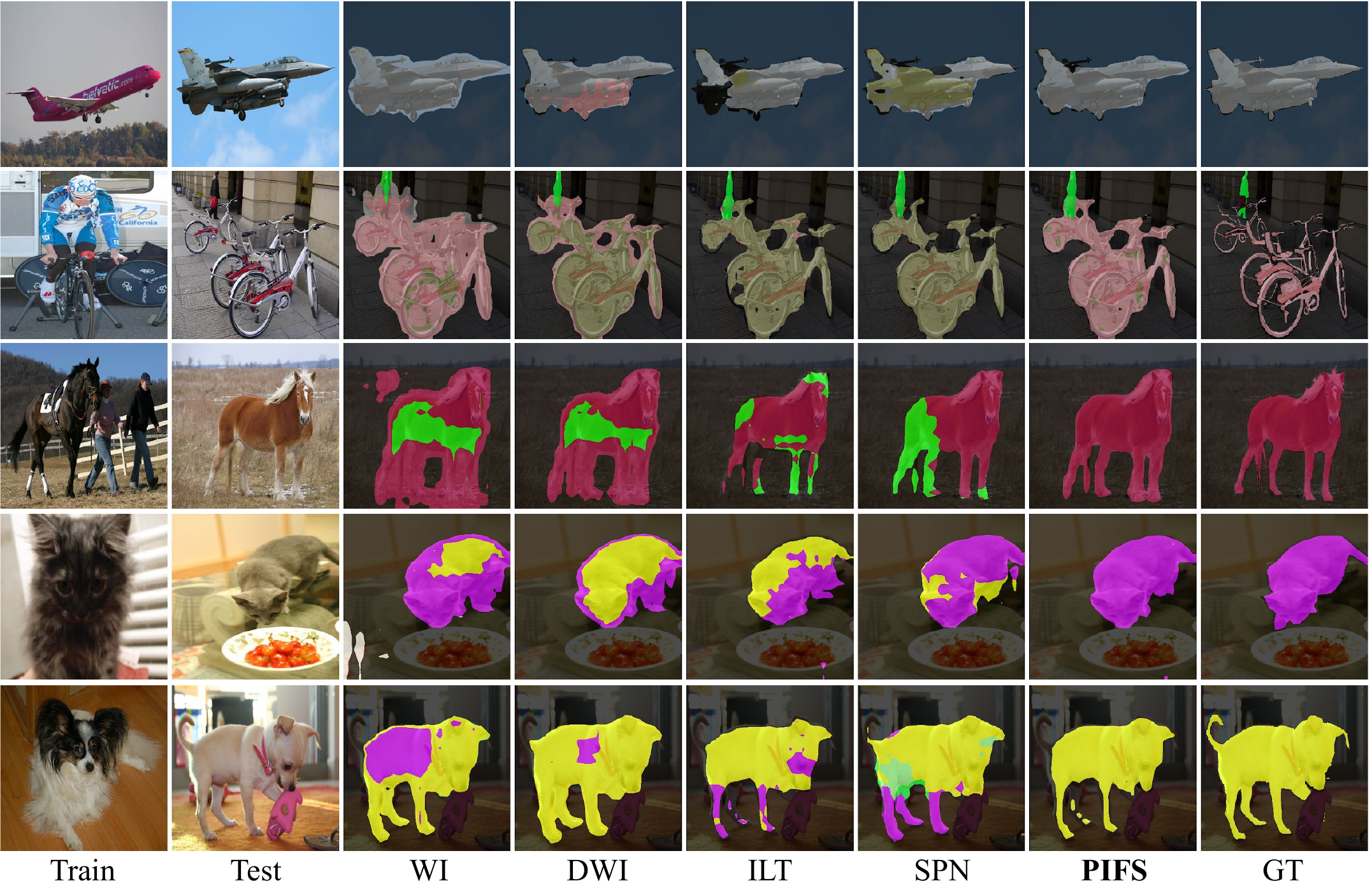}
    \vspace{-8pt}
    \caption{Qualitative results on the VOC-SS 1-shot setting.}
    \label{fig:qualitative-ablation}
    \vspace{-10pt}
\end{figure*}

\subsection{\SET: Single few-shot learning step}% addition \zeynep{this is confusing} of new classes}
%\zeynep{is there no standard experimental protocol here with a concrete name? I think you mean the standard few-shot learning setting?}
We start our analysis with a single few-shot learning (FSL) step of 5 classes on VOC-SS and of 20 classes on COCO-SS. %, reporting the results in \tabref{tab:results-single}.%From the table, we can derive important observations. 
As shown in \tabref{tab:results-single}, \ours\ achieves the top results on every dataset and shot. %In detail, \ours\ is the best in the new classes in all scenarios, while being comparable to the best results in the base ones. 
As a result, \ours\ outperforms on average the \textit{best} IL method by 3.2\% and 5.6\% in HM, and the \textit{best} FSL one by 6\% and 2.6\%, on VOC-SS and COCO-SS respectively. SPN \cite{xian2019spnet} achieves similar performance on 2 and 5 shot settings (+0.1 HM on VOC-SS 2-shot), but uses word embeddings to improve generalization on new classes. Despite not using any external knowledge, \ours\ outperforms SPN with margin on the 1-shot settings, achieving +2.8\% HM on VOC-SS, and +2\% HM on COCO-SS.
\rev{We note that some methods (\eg, DWI, ILT) surpass PIFS on the mIoU-B metric. However, they achieve sub-optimal results on new classes, either because of freezed representations  (e.g. DWI) or not exploit prototype-learning (e.g. ILT). At the cost of a slight decrease in mIoU-B, PIFS achieves the best results on new classes and the best trade-off between learning and remembering. %However, PIFS shows the best trade-off between learning the new classes and remembering the past (as shown by the HM metric, where \massi{maybe report some explanations}). %Other methods (\eg DWI, RT, ILT) achieve better results on mIoU-B but they do not achieve high results on new classes, obtaining a poor trade-off.
}

\figref{fig:qualitative-ablation} shows qualitative results for different methods on VOC-SS 1-shot. As the figure shows, WI and DWI, with fixed representations, either focus %too much 
on the context (\eg \textit{horse}, third row) or assign pixels to related classes (\eg \textit{bicycle} vs \textit{motorbike}, second row), a problem shared with ILT and SPN (\eg \textit{dog}, last row). Instead, \ours\, %learning more robust representations for new classes 
provides precise segmentation masks even when the train sample significantly differs from the test one (\eg \textit{cat}, last row). 

\begin{table*}[t]
    \centering
    \setlength{\tabcolsep}{3pt} % Default value: 6pt
    
    \resizebox{\linewidth}{!}
    {\begin{tabular}{ll|ccc|ccc|ccc||ccc|ccc|ccc}
    & & \multicolumn{9}{c||}{\textbf{VOC-MS}} & \multicolumn{9}{c}{\textbf{COCO-MS}}  \\
    & &  \multicolumn{3}{c}{\textbf{1-shot}} & \multicolumn{3}{c}{\textbf{2-shot}} & \multicolumn{3}{c||}{\textbf{5-shot}} & \multicolumn{3}{c}{\textbf{1-shot}} & \multicolumn{3}{c}{\textbf{2-shot}} & \multicolumn{3}{c}{\textbf{5-shot}} \\ \hline
     & Method	& mIoU-B & mIoU-N	 & HM	 & mIoU-B & mIoU-N	 & HM & mIoU-B & mIoU-N	 & HM & mIoU-B & mIoU-N	 & HM	 & mIoU-B & mIoU-N	 & HM & mIoU-B & mIoU-N	 & HM  \\ \hline
& FT                                                  & 47.2&	3.9&	7.2&	53.5&	4.4&	8.1&	58.7&	7.7&	13.6   & 38.5&	4.8&	8.5&	40.3&	6.8&	11.6&	39.5&	11.5&	17.8       \\ \hline
\parbox[t]{2mm}{\multirow{3}{*}{\rotatebox[origin=c]{90}{FSC}}}
& WI \cite{qi2018low}                       & {66.6}&	16.1&	25.9&	66.6&	19.8&	30.5&	66.6&	21.9&	33.0   & \textbf{46.3}&	8.3&	14.1&	\textbf{46.5}&	9.3&	15.5&	46.3&	10.3&	16.9       \\
& DWI \cite{gidaris2018dynamic}   & \textbf{67.2}&	16.3&	26.2&	\textbf{67.5}&	21.6&	32.7&	\textbf{67.6}&	25.4&	36.9   & 46.2&	9.2&	15.3&	\textbf{46.5}&	11.4&	18.3&	\textbf{46.6}&	14.5&	22.1       \\
& RT \cite{tian2020rethinking}         & 49.2&	5.8&	10.4&	36.0&	4.9&	8.6&	45.1&	10.0&	16.4   & 38.4&	5.2&	9.2&	43.8&	10.1&	16.4&	44.1&	16.0&	23.5       \\ \hline
\parbox[t]{2mm}{\multirow{2}{*}{\rotatebox[origin=c]{90}{FSS}}}
& AMP  \cite{siam2019adaptive}          & 58.6&	14.5&	23.2&	58.4&	16.3&	25.5&	57.1&	17.2&	26.4   & 36.6&	7.9&	13.0&	36.0&	9.2&	14.7&	33.2&	11.0&	16.5       \\ 
& SPN  \cite{xian2019spnet}              & 49.8&	8.1&	13.9&	56.4&	10.4&	17.6&	61.6&	16.3&	25.8   & 40.3&	8.7&	14.3&	41.7&	12.5&	19.2&	41.4&	18.2&	\textbf{25.3} \\ \hline 
\parbox[t]{2mm}{\multirow{3}{*}{\rotatebox[origin=c]{90}{IL}}}

& LwF  \cite{li2017learning}                & 42.1 &	3.3 &	6.2 &	51.6 &	3.9 &	7.3 &	59.8 &	7.5 &	13.4    & 41.0&	4.1&	7.5&	42.7&	6.5&	11.3&	42.3&	12.6&	19.4       \\
& ILT  \cite{michieli2019incremental}       & 43.7 &	3.3 &	6.1 &	52.2 &	4.4 &	8.1 &	59.0 &	7.9 &	13.9    & 43.7&	6.2&	10.9&	{47.1}&	{10.0}&	{16.5}&	45.3&	15.3&	22.9       \\
& MiB  \cite{cermelli2020modeling}          & 43.9 &	2.6 &	4.9 &	51.9 &	2.1 &	4.0 &	60.9 &	5.8 &	10.5    & 40.4&	3.1&	5.8&	42.7&	5.2&	9.3&	43.8&	11.5&	18.2       \\ \hline
%\ours	    & 63.8&	16.3&	26.0&	64.5&	23.4&	34.3&	63.7&	27.2&	38.1   & 42.4&	10.7&	17.1&	41.7&	13.6&	20.5&	42.7&	19.7&	27.0       \\
%WIT         & 43.4&	9.0&	14.9&	52.5&	13.1&	21.0&	54.8&	20.1&	29.4   & 37.3&	8.9&	14.4&	38.0&	12.1&	18.4&	38.9&	18.0&	24.6       \\ \hline
& \textbf{\ours}     & 64.1&	\textbf{16.9}&	\textbf{26.7}&	65.2&	\textbf{23.7}&	\textbf{34.8}&	64.5&	\textbf{27.5}&	\textbf{38.6}   & 40.4&	\textbf{10.4}&	\textbf{16.5}&	40.1 &	\textbf{13.1} &	\textbf{19.8} &	41.1&	\textbf{18.3}&	\textbf{25.3} \\ 
    \end{tabular}}
    \vspace{-2pt} \caption{\SET: average mIoU across steps on multi few-shot learning step scenarios.} \label{tab:results-multi}
    \vspace{-15pt}
\end{table*}

\comment{
\begin{figure}
  \centering
\begin{subfigure}{0.3\textwidth}
  \centering
  \includegraphics[width=\linewidth]{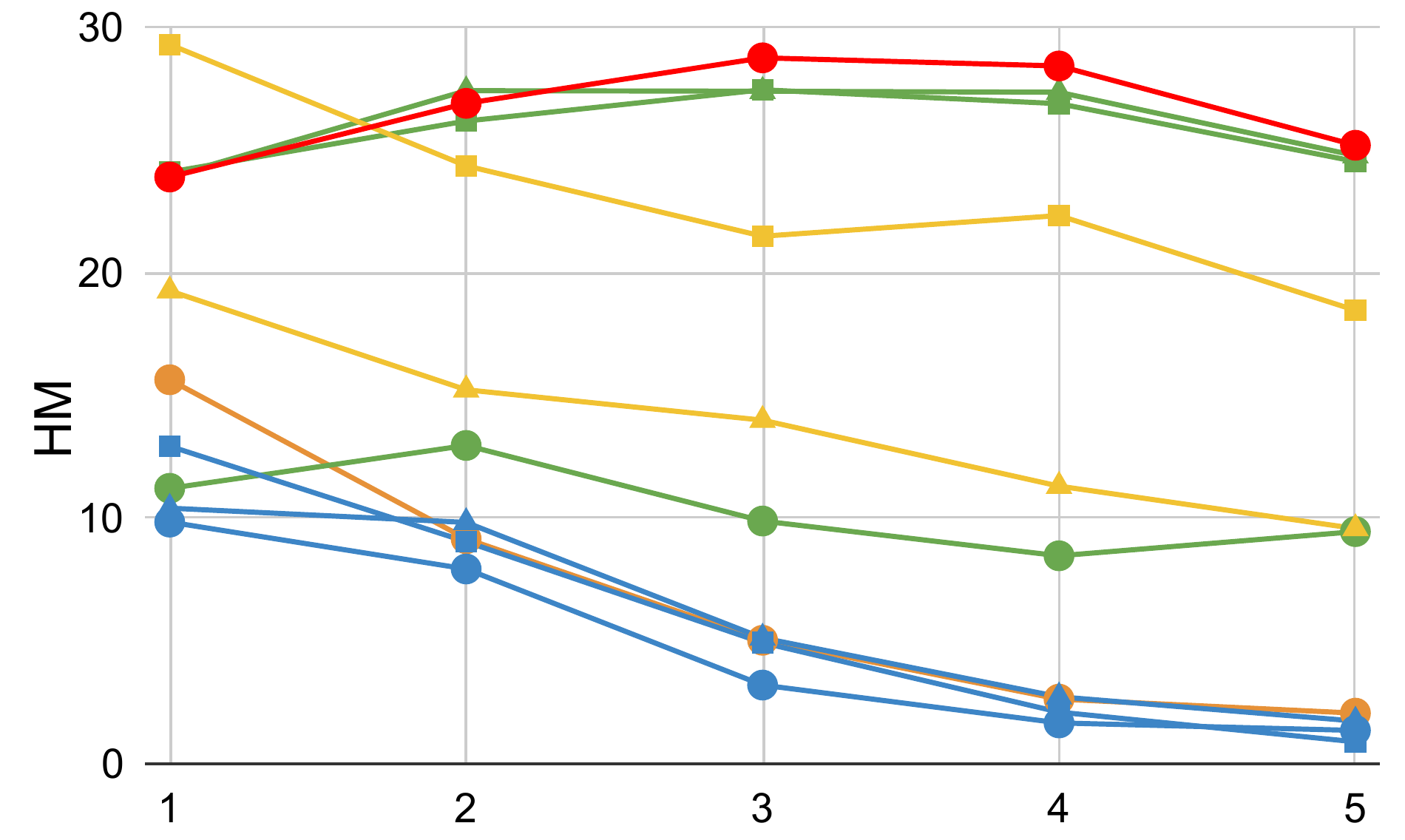}
  \caption{VOC 1-shot}
  \vspace{-5pt}
\end{subfigure}%
\begin{subfigure}{0.3\textwidth}
  \centering
  \includegraphics[width=\textwidth]{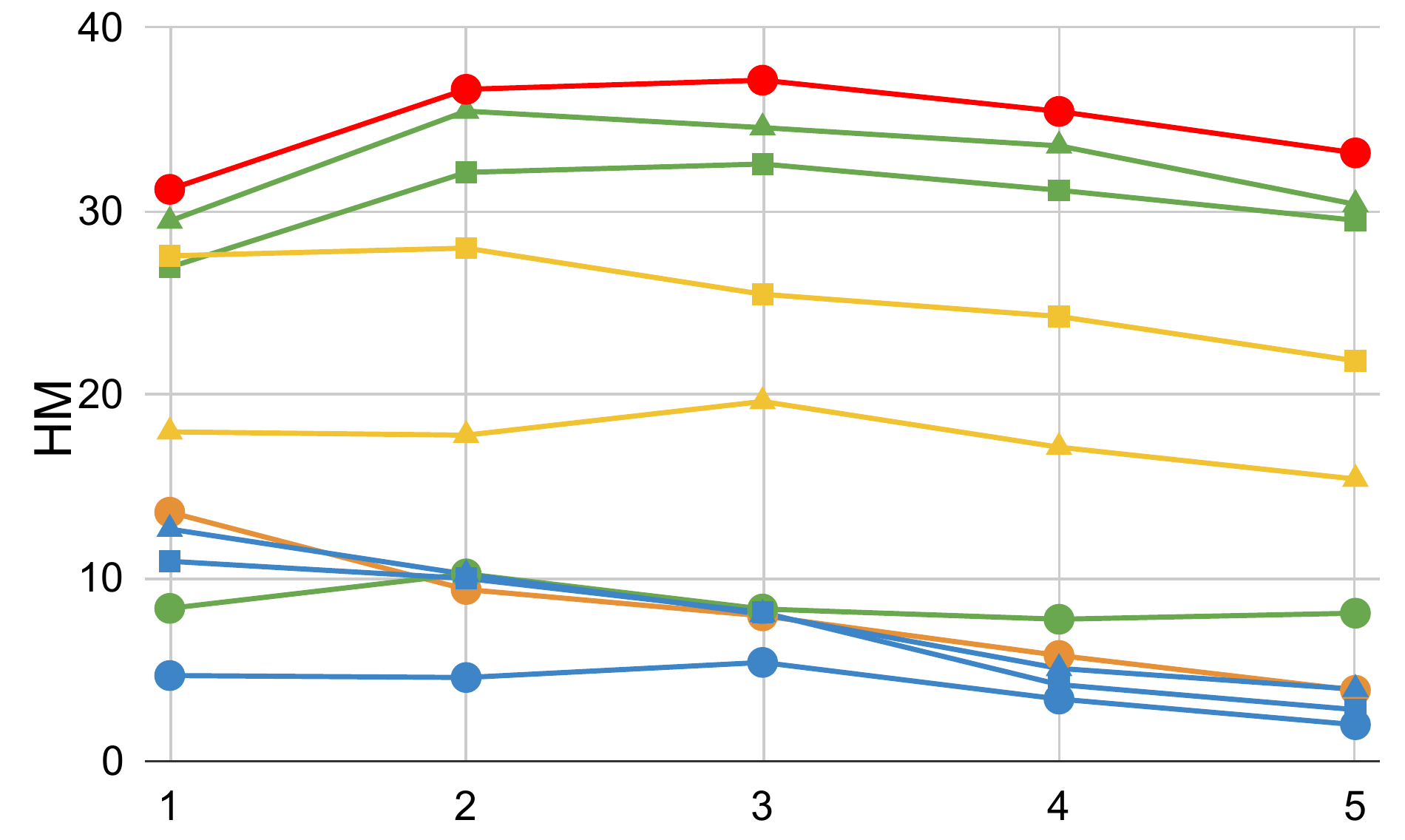}
  \caption{VOC 2-shot}
    \vspace{-5pt}
\end{subfigure}%
\begin{subfigure}{0.3\textwidth}
  \centering
  \includegraphics[width=\textwidth]{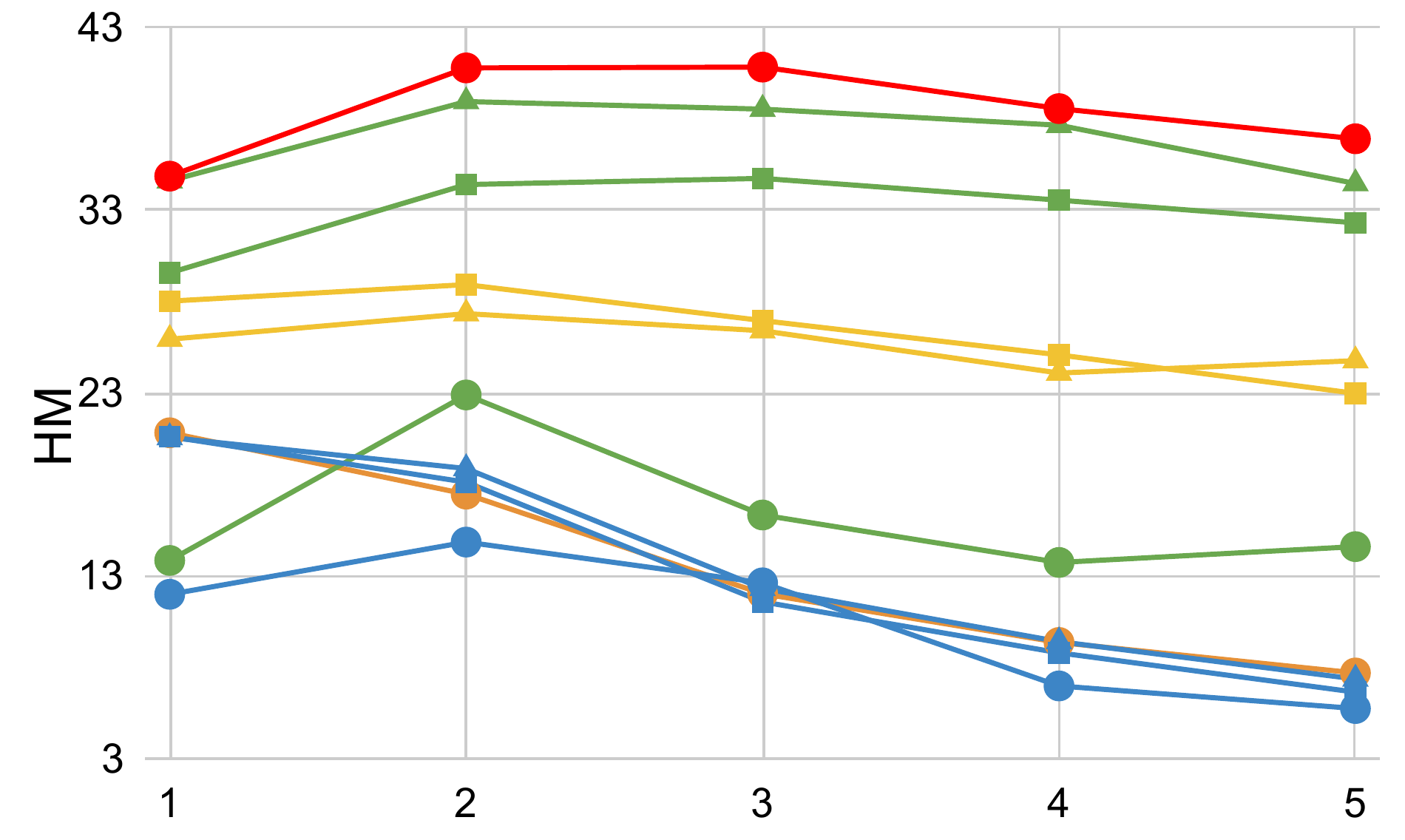}
  \caption{VOC 5-shot}
    \vspace{-5pt}
\end{subfigure}
\begin{subfigure}{0.056\textwidth}
  \centering
  \includegraphics[width=\textwidth]{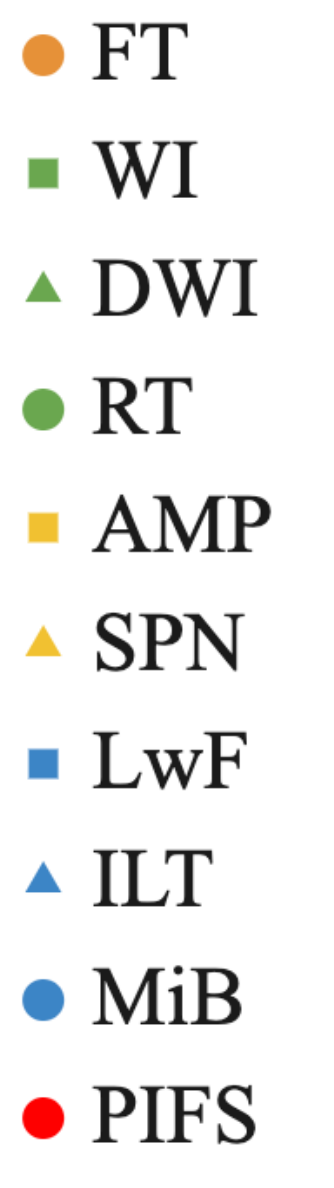}
    \vspace{2pt}
\end{subfigure}

\begin{subfigure}{0.3\textwidth}
  \centering
  \includegraphics[width=\textwidth]{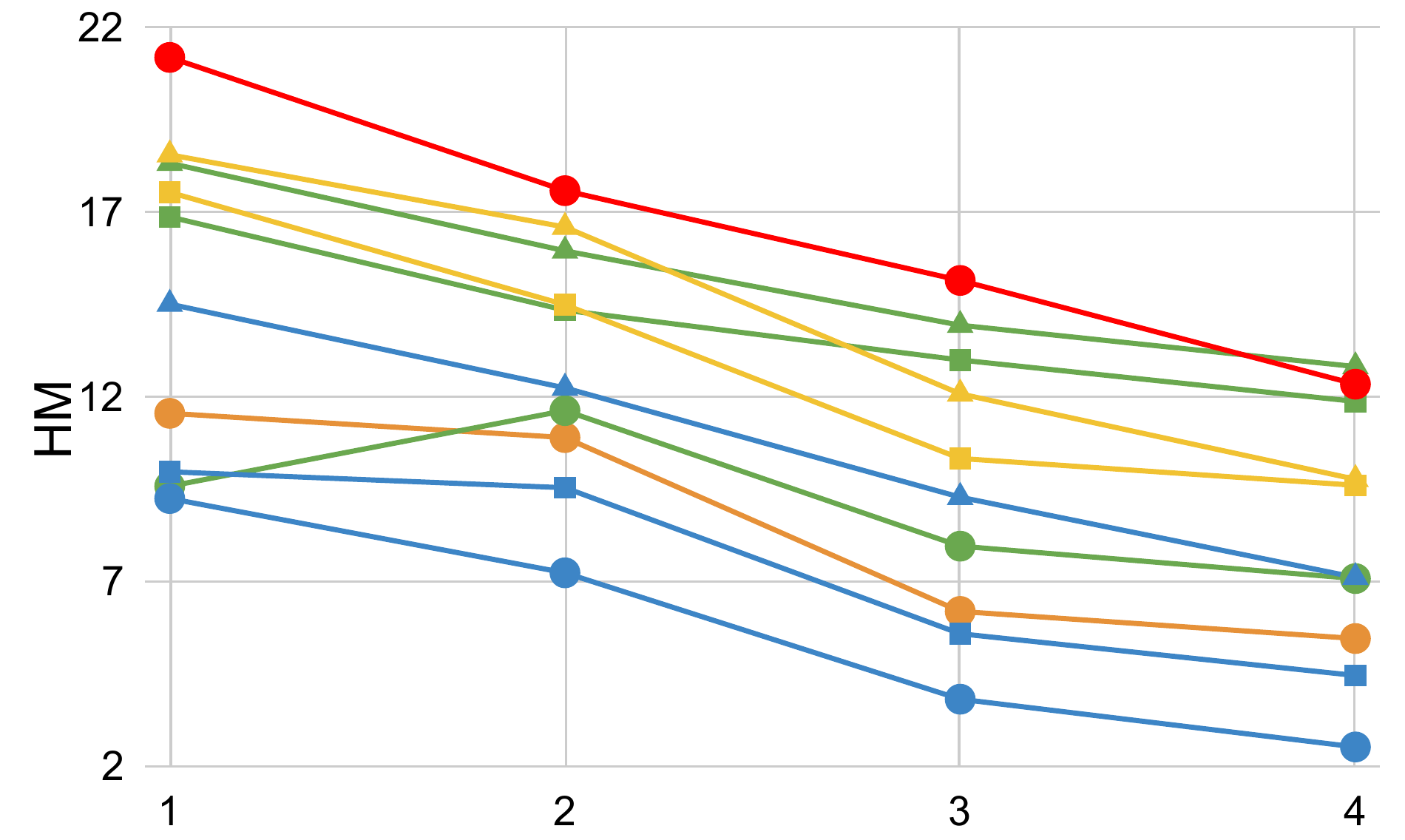}
  \caption{COCO 1-shot}
\end{subfigure}
\begin{subfigure}{0.3\textwidth}
  \centering
  \includegraphics[width=\textwidth]{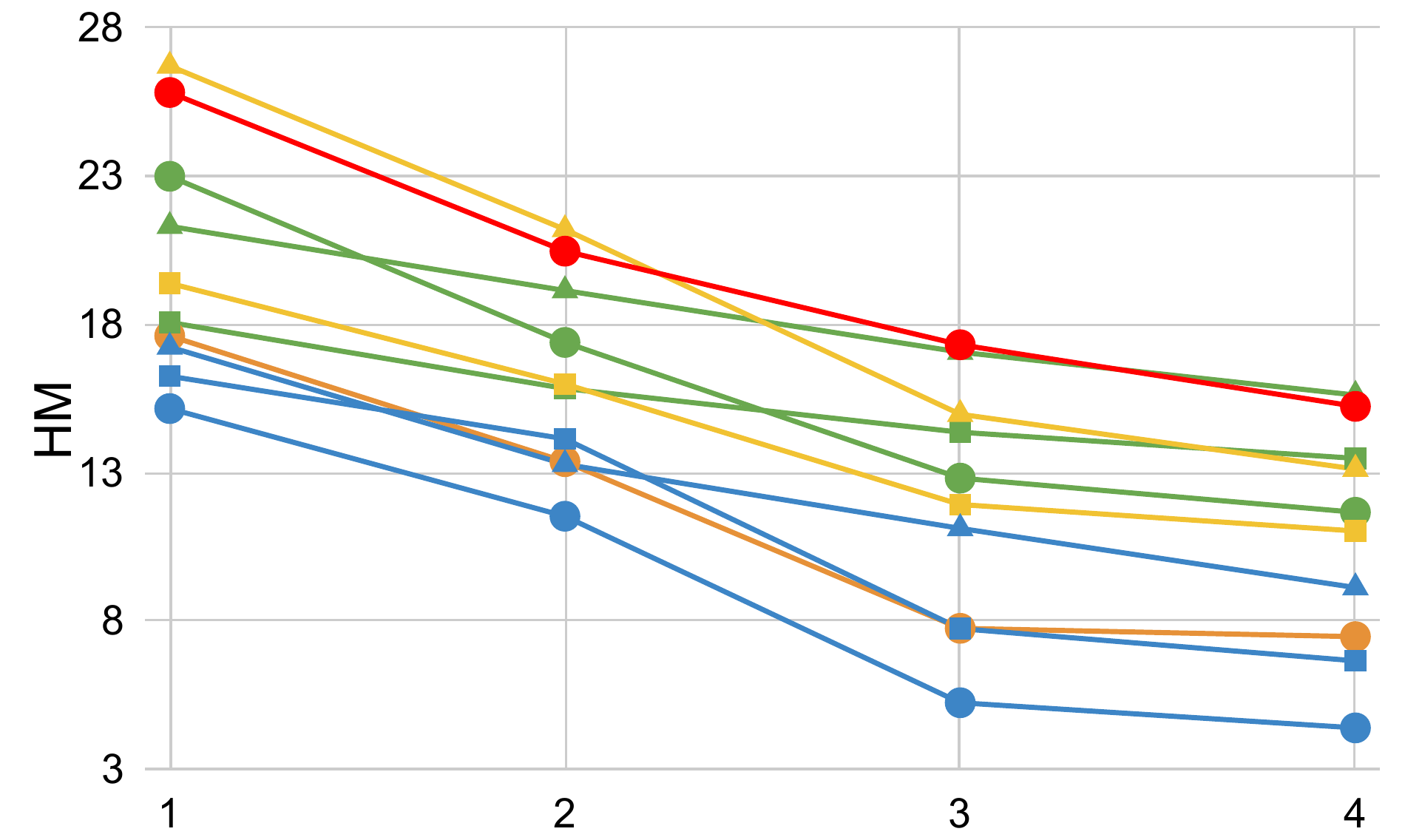}
  \caption{COCO 2-shot}
\end{subfigure}
\begin{subfigure}{0.3\textwidth}
  \centering
  \includegraphics[width=\textwidth]{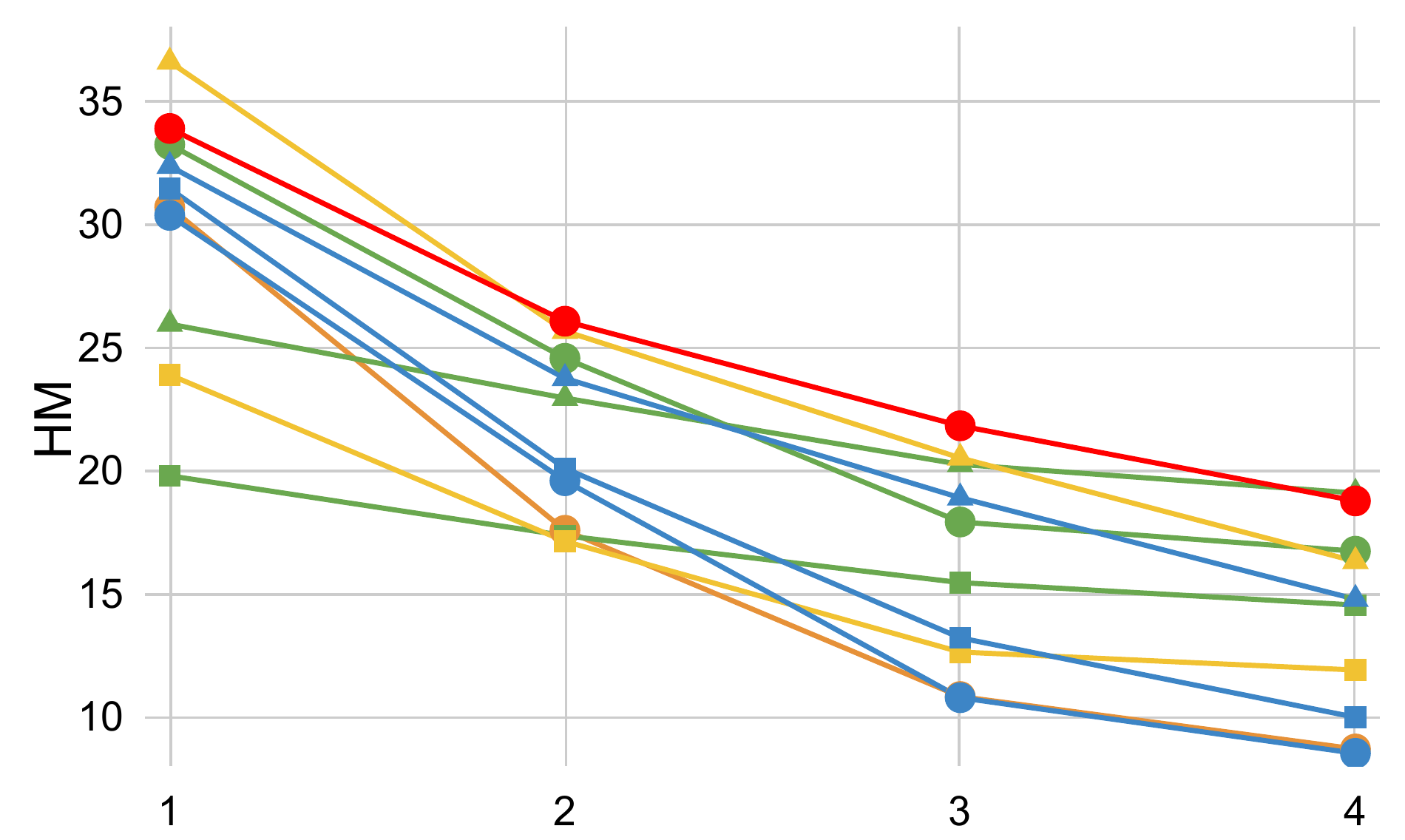}
  \caption{COCO 5-shot}
\end{subfigure}
\begin{subfigure}{0.056\textwidth}
  \centering
  \includegraphics[width=\textwidth]{fig/plot22/legend.pdf}
  \vspace{2pt}
\end{subfigure}
\vspace{2pt}
\caption{\SET\ results on the sequential addition of new class. Every column is a new step.}
\vspace{-10pt}
\label{fig:multi-step}
\end{figure}
}

\subsection{\SET: Multiple few-shot learning steps}
%While the single FSL step scenarios are interesting to assess the ability to learn multiple new classes, they do not stress the ability of a model to alleviate catastrophic forgetting.
%For this reason, we design an additional set of experiments where
In this section, we test all methods under multiple FSL steps, \ie 5 steps of 1 class (VOC-MS) and 4 steps of 5 classes (COCO-MS). Note that VOC-MS is challenging due to the scarce number of training images, \ie as little as one in the 1-shot case. \rev{We report the average performance obtained in the multiple FSL steps on Tab.~\ref{tab:results-multi}}. Step-by-step results are reported in the supplementary material.

\ours\ yields a significant improvement over the baselines, outperforming on average in HM the {best} IL method by 12.9\% and 5.0\%, and the best FSS one by 8.3\% and 0.9\% %, and the FT baseline by 23.7\% and 7.9\%,
on VOC-MS and COCO-MS respectively. 
% Note that while on base classes it performs comparably with top methods, on new ones 
Note that \ours\ is always superior on new classes, outperforming the best non end-to-end method (WI, DWI, AMP, RT) by 2\% on average, demonstrating the benefits of fine-tuning even in the extreme scenario with only one training image.
%On the other hand, 
FT instead fails on this scenario %(\eg 7.2\% HM on VOC-MS, 8.5\% HM on COCO-MS), 
showing that our prototype learning and the distillation loss are crucial to avoid forgetting old classes and overfitting on new ones. 

For what concerns IL methods, they struggle on learning new classes, improving over FT on COCO-MS 2 and 5 shots thanks to their knowledge distillation losses (\ie +1.2\% HM on 2-shot, +5.1\% on 5-shot for ILT). However, they are still far from \ours\, \ie -7\% and -2.4\% HM on COCO-MS 2-shot and 5-shot respectively. In VOC-MS and in the 1-shot settings, the comparison is even more evident, with \ours\ outperforming the best IL method of 20.5\% HM on VOC-MS and 26.7\% HM on COCO-MS. This is because it is extremely hard to learn new classes from scratch on few images without exploiting prototype learning.

Finally, SPN suffers forgetting when learning from tiny datasets (i.e. VOC-MS 1-shot) where
\ours\ still outperforms it (i.e. +12.8\% HM) despite no use of external knowledge. 
%Using  allows \ours\ to avoid both %to maintain the drift on the statistics and overfitting on the non i.i.d. few-shot data. % whisame training statistics on the FSL, while SPN suffers a drift, leading to catastrophic forgetting.

% We note that increasing the number of shots is highly beneficial for all the methods. We show this on the Pascal-VOC dataset, where the performance increases of 5\% both going from 1-shot to 2-shot, and from 2-shot to 5-shot. 
% We note that some classes are harder to learn than others, \eg \textit{potted-plant} and \textit{tv-monitor}, for which the performance are poor. This is probably because there are no similar classes on the base step, making it hard for the network to transfer its knowledge to new classes. However, on these classes it is highly beneficial to train our network, as indicated by the gap among WI and \ours, evident in the 5-shot case.

\begin{table}[t]
    \centering
    \setlength{\tabcolsep}{10pt} % Default value: 6pt
    \resizebox{0.75\linewidth}{!}{
    \begin{tabular}{cccc|ccc|ccc}
    
    \multicolumn{4}{c|}{}  &  \multicolumn{3}{c|}{\textbf{VOC-SS 1-shot}} & \multicolumn{3}{c}{\textbf{COCO-SS 1-shot}}  \\ \hline
FT    &  WI    & BR & $\ell_{KD}$ & mIoU-B & mIoU-N & HM & mIoU-B & mIoU-N	 & HM  \\ \hline
\cmark &        &    &                             & 58.3 &	 9.7 &	16.7 & 41.2 &	4.1 & 7.5 \\ % FT
\cmark &        &    & KD       & 61.5 &	10.7 &	18.2 & \textbf{43.9} &	3.8 & 7.0 \\ % KD
\cmark &        &    & L2                           & 61.3 &	10.4 &	17.8 & 43.3 & 3.3 & 6.1 \\ % L2
       & \cmark & &                               & \textbf{62.7}&	15.5 &	24.8 & 43.8 &	6.9 &	11.9 \\ % WI
\cmark & \cmark & &                                & 56.6 &	14.0 &	22.5 & 39.9 &	7.4 &	12.5 \\ % WIT
\cmark & \cmark & & PD                              & 57.6 &	14.7 &	23.4 & 40.5 &	7.9 &	13.2 \\ % WIT + KDWarm
\cmark & \cmark & \cmark &                         & 59.9 &	17.7 &	27.4 & 39.8 &	7.4 &	12.5 \\ % WIT + BR
\cmark & \cmark & \cmark & KD   & 62.1 &	18.2 &	28.1 & 41.6 &	7.4 &	12.6 \\ % WIT + BR + KD
\cmark & \cmark & \cmark & L2                       & 61.9 &	18.4 &	28.3 & 41.2 &	7.0 &	12.0 \\ % WIT + BR + KD
\hline 
\cmark & \cmark & \cmark & PD                       & 60.9 &	\textbf{18.6} &	\textbf{28.4} & 40.8 &	\textbf{8.2} &	\textbf{13.7} \\ % OURS
%\cmark & \cmark & \cmark & L2 \\ % WIT + BR + PD 

    \end{tabular}}
    \vspace{5pt}
     \caption{Ablation of the different component of \ours.  %WI weight imprinting, BR is batch-renormalization, PD our prototype-based distillation loss while KD \cite{li2017learning} and L2 \cite{michieli2019incremental} standard ones.
    }  \label{tab:ablation}
\vspace{-18pt}   
\end{table}

\subsection{Ablation study}
We ablate all the components of our method: i) prototype initialization (WI) vs a standard random classifier, ii) end-to-end training (FT), iii) batch-renorm (BR) in place of batch normalization, and iv) our prototype knowledge distillation (PD) compared to standard ones, \ie on old class probabilities (KD) \cite{li2017learning} and L2 on features extracted from $f^{t-1}$ and $f^{t}$.
%Since we found consistent results across shots, 
\tabref{tab:ablation} {reports} results on the challenging 1-shot benchmarks of VOC-SS and COCO-SS. 

{The results of FT, FT+KD, FT+L2 show that, starting from random weights in the classifier, performance on new classes are poor. In contrast, WI alone achieves good results by exploiting prototype learning, avoiding forgetting. When training the initialized network (FT+WI), there is a clear improvement \wrt FT alone, \ie at least +5\% in HM, but a decrease of nearly 6\% and 4\% HM \wrt WI on base classes, due to catastrophic forgetting. 
%Indeed, using our distillation loss (PD) alleviates catastrophic forgetting, but does not tackle the shift in network statistics. Aligning them using BR improves the performances, especially when only few images are available (VOC). We also note that BR and PD are complementary and we achieve the best performance on both datasets under all metrics using them together.

The table shows that both PD and BR can alleviate forgetting: PD improves results on base classes on both datasets, while BR is especially helpful when few images are available (\ie 27.4\% HM on VOC). Furthermore, they are complementary and when applied together we achieve the best performance on both datasets (\eg 13.7\% HM on COCO). %, with +1\% in HM. 

Finally, we compare our distillation loss (PD) with the KD and L2 loss. While coupling them with WI largely improves the performance, our PD loss still outperforms both of them (\eg +1.7\% HM over L2 and +1.1\% HM over KD on COCO), demonstrating that is important to design a distillation loss that also reduces overfitting of new class prototypes.
%Coupling KD and L2 with prototype learning, both show large boosts in performance on new classes (\ie KD 7.5\% on VOC and 3.6\% on COCO, L2 {8.0}\% on VOC and {3.7}\% on COCO). This confirms the fundamental role that prototype learning plays when fine-tuning the whole architecture in \SET. Despite these improvements, our PD loss still outperforms both of them (\eg +1.7\% HM over L2 and +1.1\% HM over KD on COCO), demonstrating that is important to design a distillation loss that also reduces overfitting of new class prototypes.
}

\begin{table*}[t]
    \centering
    \setlength{\tabcolsep}{3pt} % Default value: 6pt
    
    \resizebox{0.9\linewidth}{!}
    {\begin{tabular}{ll|ccc|ccc|ccc||ccc|ccc|ccc}
    & & \multicolumn{9}{c||}{\textbf{VOC-SS-strict}} & \multicolumn{9}{c}{\textbf{COCO-SS-strict}}  \\
    & &  \multicolumn{3}{c}{\textbf{1-shot}} & \multicolumn{3}{c}{\textbf{2-shot}} & \multicolumn{3}{c||}{\textbf{5-shot}} & \multicolumn{3}{c}{\textbf{1-shot}} & \multicolumn{3}{c}{\textbf{2-shot}} & \multicolumn{3}{c}{\textbf{5-shot}} \\ \hline
     & Method	& mIoU-B & mIoU-N	 & HM	 & mIoU-B & mIoU-N	 & HM & mIoU-B & mIoU-N	 & HM & mIoU-B & mIoU-N	 & HM	 & mIoU-B & mIoU-N	 & HM & mIoU-B & mIoU-N	 & HM  \\ \hline
& FT                                        & 55.0 & 10.2 & 17.2 & 55.5 & 19.2 & 28.6 & 43.7 & 26.8 & 33.2 &    35.3 &  4.5 &  8.0 & 32.8 &  7.4 & 12.1 & 26.9 & 11.1 &	15.7 \\ \hline
\parbox[t]{2mm}{\multirow{3}{*}{\rotatebox[origin=c]{90}{FSC}}}
& WI \cite{qi2018low}                       & 62.7 & 15.5 & 24.8 & 63.3 & 19.2 & 29.5 & 63.3 & 21.7 & 32.3 &    43.8 &  6.9 & 11.9 & 44.2 &  7.9 & 13.5 & 43.6 &  8.7 &	14.6 \\
& DWI \cite{gidaris2018dynamic}             & \textbf{64.3} & 15.4 & 24.8 & \textbf{64.8} & 19.8 & 30.4 & {64.9} & 23.5 & 34.5 &    44.5 &  7.5 & 12.8 & 45.0 &  9.4 & 15.6 & 44.9 & 12.1 &	19.1 \\
& RT \cite{tian2020rethinking}              & 60.1 & 11.0 & 18.6 & 62.3 & 19.7 & 29.9 & 61.0 & 26.0 & 36.5 &    \textbf{46.0} &	4.0 & 7.3  & \textbf{46.5} &	 5.1 &  9.2	& \textbf{46.8} &  7.5 & 13.0 \\ \hline
\parbox[t]{2mm}{\multirow{2}{*}{\rotatebox[origin=c]{90}{FSS}}}
& AMP  \cite{siam2019adaptive}              & 56.6 & 16.6 & 25.7 & 54.6 & 18.8 & 28.0 & 51.6 & 18.2 & 26.9 &    42.7 &  6.8 & 11.8 & 42.7 &  8.2 & 13.7 & 42.4 & 10.0 &	16.2 \\ 
& SPN  \cite{xian2019spnet}                 & 56.4 & 16.4 & 25.4 & 57.1 & 25.3 & 35.1 & 48.7 & 30.2 & 37.3 &    38.1 &  7.0 & 11.8 & 37.0 & 10.4 & 16.3 & 33.2 & 15.1 &	20.8 \\ \hline 
\parbox[t]{2mm}{\multirow{3}{*}{\rotatebox[origin=c]{90}{IL}}}
& LwF  \cite{li2017learning}                & 60.6 & 11.2 & 18.9 & 62.8 & 19.5 & 29.8 & 56.2 & 29.7 & 38.9 &    43.0 &  4.5 &  8.1 & 42.6 &  8.3 & 13.9 & 40.6 & 13.7 &	20.5 \\
& ILT  \cite{michieli2019incremental}       & 63.1 & 14.1 & 23.0 & 63.6 & 23.8 & 34.7 & 58.9 & 31.6 & 41.2 &    45.2 &  5.1 &  9.2 & 45.0 &  8.0 & 13.6 & 44.0 & 13.3 & 20.4 \\
& MiB  \cite{cermelli2020modeling}          & 61.0 &  6.1 & 11.1 & 63.6 & 13.7 & 22.6 & \textbf{65.0} & 29.4 & 40.5 &    43.7 &  4.2 &  7.7 & 44.2 &  7.1 & 12.3 & 44.4 & 13.8 &	21.1 \\ \hline
& \textbf{\ours}                            & 59.1 & \textbf{18.3} & \textbf{27.9} & 58.8 & 26.2 & 36.2 & 57.2 & 32.6 & 41.5 &    34.9 &  \textbf{8.9} & 14.2 & 34.6 & 11.7 & 17.4 & 32.6 & 15.6 &	21.1  \\ 
& \textbf{\ours*}                           & 60.3 & 18.0 & 27.8 & 60.3 & \textbf{26.3} & \textbf{36.6} & 59.6 & \textbf{33.1} & \textbf{42.5} &    38.8 &  8.8 & \textbf{14.4} & 39.2 &\textbf{ 11.8} & \textbf{18.1} & 38.4 & \textbf{16.1} &	\textbf{22.6}  \\ 
    \end{tabular}}
    \vspace{3pt}
    \caption{\rev{\SET: mIoU on single few-shot learning step scenarios with background shift. \ours* uses the revised cross-entropy loss of \cite{cermelli2020modeling}}.} \label{tab:results-strict}
    \vspace{-16pt}
\end{table*}

\subsection{\SET\ with background shift}
\rev{
In the previous settings, we assumed that old class pixels were annotated in the FSL steps. {However this assumption might be not feasible in some scenarios and in this section, following recent IL works \cite{cermelli2020modeling,douillard2021plop, michieli2021continual}, we annotate the old class pixels as background.} %Differently, in this section, following \cite{cermelli2020modeling, douillard2021plop, michieli2021continual}
%we annotate the old classes' pixels as background, as in recent IL works \cite{cermelli2020modeling,douillard2021plop, michieli2021continual}. 
Note that, in such scenario, we introduce the background shift problem \cite{cermelli2020modeling}, i.e. the semantic of the background changes across incremental steps and may contain old classes, exacerbating catastrophic forgetting. To test this setting, we adhere to the \textit{disjoint} protocol of \cite{cermelli2020modeling}, excluding from the base step dataset all the images containing pixels from new classes.

Table~\ref{tab:results-strict} reports the results on the single-step settings of VOC (VOC-SS-strict) and COCO (COCO-SS-strict), considering 1, 2 or 5 images in the FSL steps. %Since the losses in cross-entropy loss of MiB and the regularization term in PIFS are orthogoanl, %
In this setting, 
we introduce PIFS* that uses the revised cross-entropy loss proposed by \cite{cermelli2020modeling} to address the background-shift. %Overall, we note that training the whole architecture on the few images is more challenging in this scenario since the background-shift exacerbates catastrophic forgetting. 
Overall, we see that PIFS and PIFS* obtain the best trade-off, achieving the highest HM on every setting. 
In particular, PIFS* outperforms on average in HM the \textit{best} IL method by 2.7\% and 4\%, and the best FSL one by 5.7\% and 2.5\%, on VOC and COCO respectively.
SPN fails to model the background shift, obtaining poor performance on mIoU-B, with PIFS* outperforming it on average in mIoU-B by 6\% on VOC and 2.7\% on COCO.
We also note that methods that only {compute the classifiers' weights from new class pixels (i.e. WI, DWI) are not influenced by the old classes annotations and achieve the same performance to the non-strict setting (\tabref{tab:results-single}). However, PIFS shows better results, outperforming the best of them (DWI) in HM on average by 5.7\% on VOC and 2.5\% on COCO.}

Comparing PIFS and PIFS*, we note that modeling the background shift is beneficial to remember old classes, as demonstrated by the higher mIoU-B achieved by PIFS*. In particular, it outperforms on average in mIoU-B PIFS by 1.7\% on VOC and 4.7\% on COCO. However, we note that the choice of the cross-entropy loss is orthogonal to the contributions of PIFS and can be easily integrated when the background shift is present.

}

\vspace{-8pt}
\section{Conclusion}
\comment{In this work we proposed \setting\ (\SET), whose goal is to extend a pretrained segmentation model with new classes given few annotated images and without access to old training data, combining the challenges of few-shot and incremental learning. 
Additionally, we proposed the first model for \SET, \ours, which combines prototype learning with regularized fine-tuning on the new classes. While prototype learning allows for robust initialization of the classifier, fine-tuning improves the representation of new classes. To reduce catastrophic forgetting and overfitting at once, we design a distillation loss where prototypes of new classes are imprinted in the old model.
Finally, batch-renorm is used to cope with the non-\textit{i.i.d}. few-shot datasets.
{We design an extensive benchmark for \SET\ using two different datasets, different number of classes, images, and learning steps.} 
Experiments show that \ours\ outperforms multiple incremental and few-shot methods on the \SET\ benchmark. 
Both the benchmark, the baselines and \ours\ will be publicly released. We hope that our novel problem formulation, broad benchmark and effective approach will serve as base for future research.}

{
In this work, we defined and studied \SET, whose goal is to extend a pretrained segmentation model with new classes given few annotated images and without access to old training data, combining the challenges of few-shot and incremental learning. 
To overcome the limitations of standard methods in \SET, we propose \ours, a method that unifies prototype learning with knowledge distillation, to achieve robust initialization of the parameters for the classifier on new classes and improve the network features representation. The distillation loss of \ours\ exploits prototypes of new classes as additional regularizer to avoid overfitting and forgetting at once. Moreover, we use batch-renorm in the few-shot learning steps to cope with the non-\textit{i.i.d.} few-shot datasets.
We designed an extensive benchmark for \SET, showing that \ours\ outperforms multiple incremental and few-shot methods.
We hope that our novel problem formulation, broad benchmark and effective approach will serve as base for future research.
}

\small
\subsubsection*{Acknowledgments}
Computational resources were partially provided by the Franklin cluster of the Italian Institute of Technology.

\bibliography{bib}
\end{document}